\documentclass{article}
\PassOptionsToPackage{hyphens}{url}
\usepackage{iclr2026_conference}
\usepackage[T1]{fontenc}
\usepackage[table]{xcolor}
\usepackage{graphicx}
\usepackage{tabularx}
\usepackage{amsmath}
\usepackage{amssymb}
\usepackage{natbib}
\usepackage{amsfonts}
\usepackage{booktabs}
\usepackage{nicefrac}
\usepackage{microtype}
\usepackage{algorithm}
\usepackage{algpseudocode}
\usepackage{array}
\usepackage{placeins}
\usepackage{hyperref}
\usepackage{url}
\newcommand{\meanstd}[2]{#1 $\pm$ #2}

\newcolumntype{L}[1]{>{\raggedright\arraybackslash}p{#1}}
\usepackage{wrapfig,lipsum,booktabs}

\title{NewtonGS: Physics-Structured Object-Level Neural Newtonian Dynamics for Gaussian Scene Animation}
\author{
Lianlei Shan \quad
Feiyang Ye\thanks{Corresponding author} \quad
Yan Chen \quad
Yong Wu\thanks{Project lead}\
\
}

\begin{document}

\renewcommand{\thefootnote}{}
\footnotetext[0]{Email: \textit{\{shanlianlei,yefeiyang,chenyan12,wuyong5\}@lixiang.com.}}
\maketitle
\fancyhead{}

\begin{abstract}
Animating objects in a static 3D Gaussian scene requires an explicit object-level dynamic state and a controllable model of object motion. Existing dynamic Gaussian methods primarily reconstruct time-varying scenes or simulate deformation, rather than provide compact object states for direct control. To address this gap, we present NewtonGS, a physics-structured framework for object-level state rollout and Gaussian scene animation. NewtonGS represents each object with a 22-dimensional state covering pose, linear and angular velocity, anisotropic scale and its rate, mass, and contact properties. Its Gaussian Neural Newtonian Dynamics (Gaussian-NND) model combines analytic translation, quaternion kinematics, gravity, damping, and scale-restoration dynamics with learned continuous and contact residuals. A discrete event map handles floor contact. Predicted poses and scales define a shared affine transformation that updates the means and covariances of all Gaussians associated with each object. We construct two procedurally generated datasets: State-32 for state-rollout evaluation and Gaussian-32 for state-to-Gaussian transformation. On both the in-distribution and velocity-range-shift splits of State-32, NewtonGS achieves lower trajectory RMSE, final displacement error, and velocity RMSE than five analytic baselines. Experiments on Gaussian-32 further demonstrate effective conversion from predicted states to animated Gaussian objects.
\end{abstract}

\section{Introduction}

3D Gaussian Splatting (3DGS) represents a scene with explicit three-dimensional primitives and supports high-quality differentiable rendering at interactive rates~\citep{kerbl2023gs}. Recent Gaussian-based generation and reconstruction methods have made it increasingly efficient to obtain static 3D assets from textual, visual, or multi-view inputs~\citep{tang2024dreamgaussian,tang2024lgm,xu2024grm}. A static Gaussian scene provides a static description of an object's geometry and appearance. It contains neither an explicit dynamic state nor a rule that determines how the object should evolve over time. Controlling this evolution requires an explicit representation of the object's dynamic state, including its position, orientation, velocity, scale, and contact properties, together with a model that determines how these variables change over time. Such control is important for Gaussian scene animation, interactive content creation, and physics-guided motion.

Several lines of work extend Gaussians beyond a static scene. Dynamic and 4D Gaussian methods represent time-varying geometry and appearance~\citep{luiten2024dynamic,wu20244dgs,yang2024native4d}, while physics-aware approaches connect visual representations with simulation, material models, or learned motion controls~\citep{yuan2026newtongen,xie2024physgaussian,zhang2024physdreamer}. These advances provide flexible ways to reconstruct, generate, or simulate dynamic content. However, these approaches are primarily designed to reconstruct dynamic appearance or simulate deformation, rather than to propose an object-level state for controlling the motion of an existing Gaussian object. Such a state is important for applications that require predictable and editable object motion, including scene animation, interactive content creation, and physics-guided control, because it allows motion variables to be specified and evaluated directly. This motivates us to study how to represent an existing Gaussian object with an explicit dynamic state, evolve that state over time, and map the predicted motion back to the Gaussian scene.

Addressing this problem presents three main challenges. First, a large set of Gaussians must be summarized by a compact object-level state whose variables retain clear geometric and dynamical meanings. Second, the state evolution must capture both continuous motion and discrete events. Translation, rotation, and scale change evolve continuously, whereas contact can produce an instantaneous change in velocity. A purely analytic model provides useful structure but may not account for model mismatch and unmodeled effects, while an unconstrained neural predictor makes the resulting motion more difficult to control and interpret~\citep{chen2018neuralode,chen2021event}. Third, the predicted object state must update all Gaussians coherently while preserving the object's internal structure, without assigning a separate trajectory to each primitive.

We introduce NewtonGS, a structured framework for object-level state rollout in Gaussian scenes. At its core, we propose Gaussian Neural Newtonian Dynamics (Gaussian-NND), a structured hybrid dynamics model that represents each Gaussian object with a 22-dimensional state containing pose, motion, anisotropic scale, and contact-related variables. Gaussian-NND combines explicit gravity, damping, quaternion, and scale dynamics with learned continuous and contact residuals. Both residual branches are initialized to produce zero outputs, so the initial rollout follows the analytic model. Contact is handled as a discrete horizontal-floor event outside the continuous solver. At each timestamp, the predicted pose and scale define a shared transformation for all Gaussian means and covariances belonging to the object.

We evaluate NewtonGS on State-32, a synthetic benchmark comprising 32 procedural motion families. Given identical labeled initial states and timestamps, NewtonGS achieves the lowest observed trajectory RMSE, final-displacement error, and velocity RMSE among five hand-specified analytic baselines on both the in-distribution and velocity-range-shift validation splits. We further apply the predicted trajectories to Gaussian objects to illustrate the deterministic state-to-Gaussian transformation.

Our contributions are threefold: 
\begin{itemize}
    \item We introduce NewtonGS, an object-level framework that formulates Gaussian object animation as the rollout of an explicit 22-dimensional state followed by a deterministic, shared transformation of the object's Gaussian primitives.
    \item We develop Gaussian-NND, a structured hybrid dynamics model that combines continuous quaternion and scale evolution, a discrete horizontal-floor contact event, and zero-output-initialized learned residuals.
    \item We evaluate NewtonGS across 32 procedural motion families under both in-distribution and velocity-range-shift settings. NewtonGS achieves lower trajectory, endpoint, and velocity errors than five analytic baselines, and its predicted states can directly animate appearance-conditioned Gaussian objects.
\end{itemize}

\section{Related Work}

\subsection{Gaussian Generation and Dynamic Representation}

Diffusion-guided methods such as DreamFusion and Magic3D established text-driven optimization of static 3D representations~\citep{poole2022dreamfusion,lin2023magic3d}. Following the introduction of 3DGS~\citep{kerbl2023gs}, DreamGaussian, LGM, and GRM improved the efficiency of generating or reconstructing canonical Gaussian assets~\citep{tang2024dreamgaussian,tang2024lgm,xu2024grm}. These methods supply the geometry and appearance of an initial scene, but do not by themselves specify how the resulting object should move.

Dynamic 3D Gaussians and 4D Gaussian Splatting (4DGS) reconstruct time-varying scenes using persistent moving primitives, deformation fields, or native space--time representations~\citep{luiten2024dynamic,wu20244dgs,yang2024native4d}. DreamGaussian4D, Align Your Gaussians, PLA4D, STP4D, and Splat4D further combine dynamic representations with video or multi-view priors for 4D generation~\citep{ren2023dreamgaussian4d,ling2024align,miao2024pla4d,deng2025stp4d,yin2025splat4d}. These representations are well suited to flexible local deformation and appearance change. NewtonGS instead starts from an existing Gaussian object and assigns it one compact object-level state, favoring a coherent object frame and direct state-space control over independently predicted primitive motion.

\subsection{Structured and Physics-Aware Dynamics}

Neural Ordinary Differential Equations (ODEs) provide a continuous-time formulation for learned state evolution~\citep{chen2018neuralode}, while Hamiltonian and Lagrangian neural networks incorporate mechanics into their parameterization~\citep{greydanus2019hamiltonian,cranmer2020lagrangian}. Neural event models extend continuous dynamics with discrete state transitions~\citep{chen2021event}. Most directly related, NewtonGen introduces Neural Newtonian Dynamics to predict an image-space physical state that guides video generation~\citep{yuan2026newtongen}. Its formulation motivates structured state evolution, although the predicted state ultimately serves as a control signal for a video generator.

Physics-aware Gaussian methods attach simulation or material models to explicit primitives. PhysGaussian couples 3D Gaussians with the Material Point Method, PhysDreamer estimates material behavior from video priors, and OmniPhysGS represents diverse constitutive responses~\citep{xie2024physgaussian,zhang2024physdreamer,lin2025omniphysgs}. They are designed for spatially varying or continuum deformation. NewtonGS targets an object-level regime: it represents translation, rotation, aggregate scale, and simple contact in a compact object-level state defined in 3D space, directly applies that state to a persistent Gaussian scene, and uses learned residuals only to correct a structured analytic update.

\section{Method}

\begin{figure*}[!ht]
\centering
\includegraphics[width=1\textwidth]{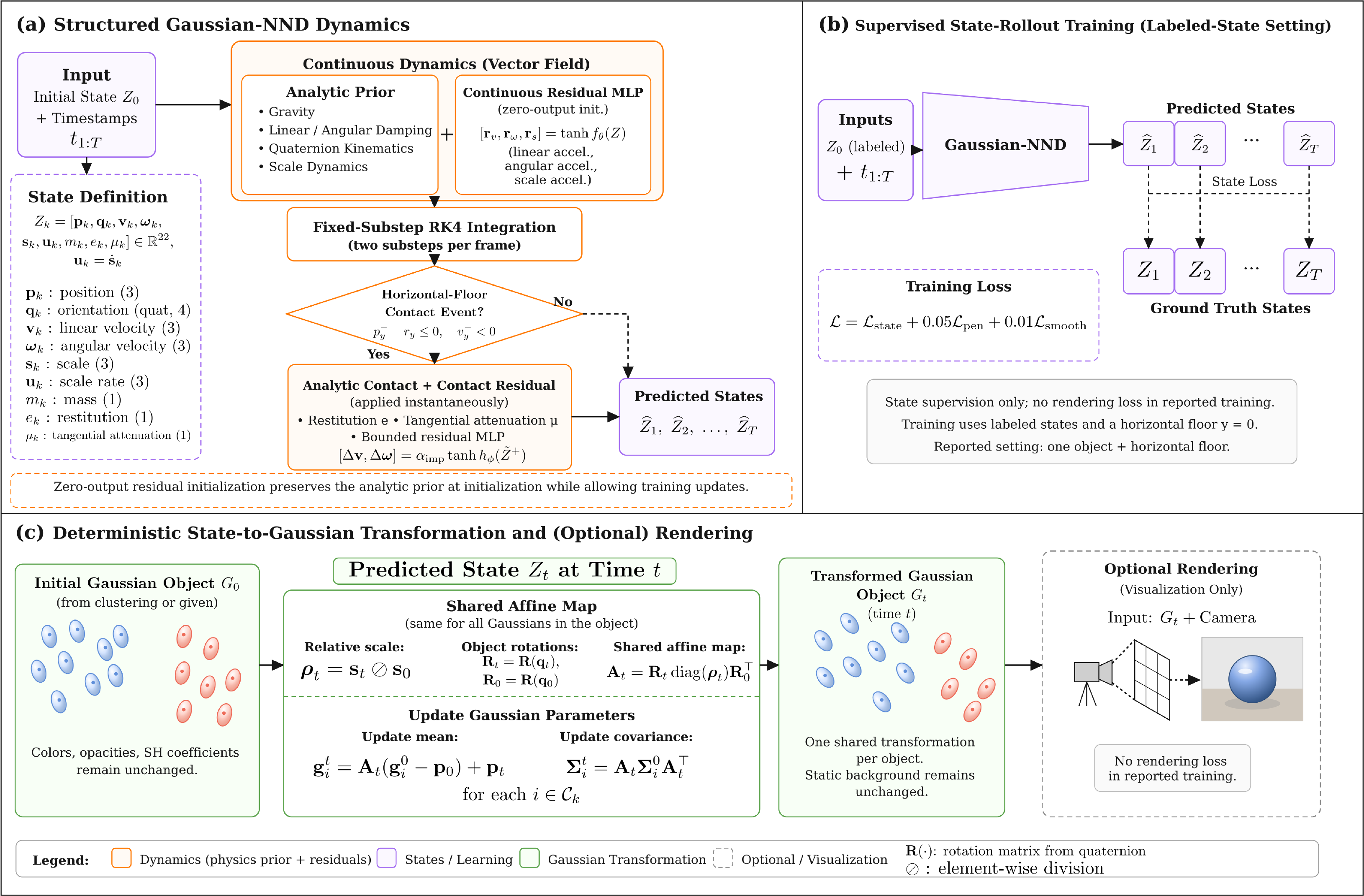}

\caption{Overview of NewtonGS. Given an object-centric Gaussian scene, we represent each dynamic object by a compact state and predict its evolution with a structured hybrid Gaussian-NND. Continuous motion is integrated by RK4, while floor contact and the nonlinear part are handled by a discrete response followed by a learned residual. The predicted pose and scale are then converted into a shared affine transformation for all Gaussians belonging to the object.}
\label{fig:method-overview}

\end{figure*}
\subsection{Overview}

We denote the initial Gaussian scene by the set $G_0=\{(\mathbf g_i^0,\boldsymbol\Sigma_i^0)\}_{i=1}^{N}$, where $\mathbf g_i^0$ and $\boldsymbol\Sigma_i^0$ denote the mean and covariance of the $i$-th Gaussian. We write only the geometric attributes needed by the dynamics; color, opacity, and other appearance attributes remain attached to each primitive. We associate each dynamic object $k$ with an index set $\mathcal C_k\subseteq\{1,\ldots,N\}$ and keep all unassigned background Gaussians static. NewtonGS first summarizes each object by a compact state $Z_0^k$, predicts its states at requested timestamps $t_{1:T}$, and maps the predicted object motion back to its Gaussians:
\begin{equation}
G_0 \rightarrow \{\mathcal C_k,Z_0^k\}_{k=1}^{K}
\rightarrow \{Z_{1:T}^k\}_{k=1}^{K}
\rightarrow G_{1:T}.
\end{equation}
As illustrated in Fig.~\ref{fig:method-overview}, the method has three main components: an object-centric Gaussian state, a hybrid dynamics model that combines an analytic prior with learned residuals, and a deterministic state-to-Gaussian transformation. Because the dynamics operate on one state per object, the network predicts one object trajectory instead of one path per primitive. The Gaussian scene remains explicit and can be rendered from any specified camera.

We distinguish the general inference pipeline from the path used to train the dynamics. At inference, object grouping together with a supplied or lifted state provides $Z_0^k$, Gaussian-NND rolls each state forward, and the resulting transformations produce $G_{1:T}$. 
In the reported state benchmark, the model takes labeled $Z_0$ as input and uses future states only for supervision. Grouping, state lifting, and rendering are not jointly optimized.

\subsection{Object-Centric Gaussian State Representation}

For each object $k$, we use the 22-dimensional state
\begin{equation}
Z_k=[\mathbf p_k,\mathbf q_k,\mathbf v_k,\boldsymbol\omega_k,
\mathbf s_k,\mathbf u_k,m_k,e_k,\mu_k],
\qquad \mathbf u_k=\dot{\mathbf s}_k,
\label{eq:state}
\end{equation}
where $\mathbf p_k\in\mathbb R^3$ and $\mathbf q_k\in\mathbb H$ are the object center and unit orientation quaternion; $\mathbf v_k,\boldsymbol\omega_k\in\mathbb R^3$ are its linear and angular velocities; and $\mathbf s_k,\mathbf u_k\in\mathbb R^3$ describe aggregate anisotropic scale and its rate. The remaining scalars are mass $m_k$, restitution $e_k$, and tangential attenuation $\mu_k$. These material channels are supplied with the initial state and remain constant during a rollout; the current model does not estimate them from appearance.

The state describes an object aggregate rather than an individual Gaussian: $\mathbf p_k$ is the cluster center, $\mathbf q_k$ specifies the aggregate axes, and $\mathbf s_k$ records scale independently of primitive covariances. A unit quaternion avoids the singularities of Euler angles~\citep{shoemake1985quaternion}. We normalize it during integration and treat $\mathbf q$ and $-\mathbf q$ as equivalent in both training and evaluation. Under the implemented Hamilton-product convention,
$\dot{\mathbf q}=\tfrac12\mathbf q\otimes[0,\boldsymbol\omega]$, so $\boldsymbol\omega$ is expressed in the corresponding body-coordinate convention. The 22-dimensional representation uses aggregate scale to cover size-changing and simple deformation trajectories; it does not introduce a separate high-dimensional deformation code.

When a Gaussian object is provided, a deterministic lifting interface obtains its pose and scale from the assigned primitives. Motion rates can be estimated by finite differences when multiple frames are available; for a single frame, they must be supplied by the user or initialized to zero. Material values are likewise supplied as metadata or set to fixed defaults. The reported supervised experiments instead read the labeled initial state directly, separating dynamics evaluation from grouping and state-estimation errors.

\subsection{Structured Hybrid Gaussian-NND}

\paragraph{Continuous dynamics.}
Between contacts, Gaussian-NND evolves the object state with a structured vector field. The analytic part encodes translation, quaternion kinematics, gravity, damping, and scale restoration. A neural residual accounts for dynamics not captured by this prior. Specifically, the continuous network produces
\begin{equation}
[\mathbf r_v,\mathbf r_\omega,\mathbf r_s]
=\tanh f_\theta(Z),
\qquad f_\theta:\mathbb R^{22}\rightarrow\mathbb R^{9},
\end{equation}
where the three outputs are bounded corrections to linear, angular, and scale acceleration, respectively. The resulting dynamics are
\begin{equation}
\begin{aligned}
\dot{\mathbf p}&=\mathbf v,\\
\dot{\mathbf q}&=\tfrac12\mathbf q\otimes[0,\boldsymbol\omega],\\
\dot{\mathbf v}&=\mathbf g-|c_v|\mathbf v/m+\alpha\mathbf r_v,\\
\dot{\boldsymbol\omega}&=-|c_\omega|\boldsymbol\omega+\alpha\mathbf r_\omega,\\
\dot{\mathbf s}&=\mathbf u,\\
\dot{\mathbf u}&=-|k_s|(\mathbf s-\mathbf 1)-|c_s|\mathbf u+\alpha\mathbf r_s,\\
\dot m&=0,\qquad \dot e=0,\qquad \dot\mu=0,
\end{aligned}
\label{eq:dynamics}
\end{equation}
where $\mathbf g=(0,-9.81,0)$, and $c_v,c_\omega,k_s,c_s$, and $\alpha$ are learned global coefficients. The residuals are acceleration corrections: $\mathbf r_v$ is not divided by mass, and $\mathbf r_\omega$ is not divided by an inertia tensor. The final layer of $f_\theta$ is initialized to zero, so the initial model follows the analytic prior while the residual is learned from data. We evaluate the complete right-hand side, including $f_\theta$, at the intermediate stages of fixed-substep RK4 integration~\citep{hairer1993ode}.

\paragraph{Contact event.}
Continuous integration alone does not represent the discontinuous velocity change at impact. We therefore use a separate event map for contact with the horizontal floor $y=0$. Let $Z^-$ be the state after a continuous substep, define the tangential velocity $\mathbf v_\parallel^-=[v_x^-,v_z^-]^\top$, and use the proxy vertical extent
$r_y=\max(|s_y^-|,10^{-3})$. A contact is detected when
\begin{equation}
p_y^- - r_y\leq 0
\qquad\text{and}\qquad
v_y^-<0.
\end{equation}
For a detected hit, the analytic response first gives
\begin{equation}
\begin{aligned}
\tilde p_y^+&=r_y,&
\tilde v_y^+&=-e v_y^-,\\
\tilde{\mathbf v}_\parallel^+&=(1-\mu)\mathbf v_\parallel^-,&
\tilde{\boldsymbol\omega}^+&=(1-0.5\mu)\boldsymbol\omega^-.
\end{aligned}
\end{equation}
All state channels not listed above are copied from $Z^-$. A contact-specific network then predicts a bounded correction,
\begin{equation}
[\Delta\mathbf v,\Delta\boldsymbol\omega]
=\alpha_{\mathrm{imp}}\tanh h_\phi(\tilde Z^+),
\qquad h_\phi:\mathbb R^{22}\rightarrow\mathbb R^{6},
\end{equation}
and sets $\mathbf v^+=\tilde{\mathbf v}^++\Delta\mathbf v$ and
$\boldsymbol\omega^+=\tilde{\boldsymbol\omega}^+
+\Delta\boldsymbol\omega$. Here $\alpha_{\mathrm{imp}}$ is a learned global residual scale. As with $f_\theta$, the output layer of $h_\phi$ is initialized to zero, and the contact residual is evaluated only for detected hits. We use distinct functions because they model different operations: $f_\theta$ corrects a continuous acceleration field, whereas $h_\phi$ corrects the instantaneous post-impact velocity. The state variable $\mu$ directly attenuates tangential and angular velocity; it is not a calibrated Coulomb-friction coefficient.

\paragraph{Hybrid rollout.}
For each requested frame interval, we divide time into fixed substeps. Each substep first applies RK4 to Eq.~\eqref{eq:dynamics}, projects the quaternion and bounded state channels back to their valid domains, and then tests the contact condition. If a hit occurs, the analytic response and $h_\phi$ are applied before the next substep. The projected state is stored at each requested timestamp. The evaluated event model treats each rollout as one object interacting with a horizontal floor; wall, slope, and pairwise-object contacts are not modeled by the current event map.

\subsection{State-to-Gaussian Transformation}

The predicted object state is converted into a shared transformation for every $i\in\mathcal C_k$. For clarity, we omit the object index below. Relative to the initial state, define
\begin{equation}
\begin{aligned}
\boldsymbol\rho_t &= \mathbf s_t\oslash\mathbf s_0,
&\qquad
\mathbf R_t &= \mathbf R(\mathbf q_t),\\
\mathbf R_0 &= \mathbf R(\mathbf q_0),
&\qquad
\mathbf A_t &= \mathbf R_t
\operatorname{diag}(\boldsymbol\rho_t)\mathbf R_0^\top,
\end{aligned}
\label{eq:object-affine}
\end{equation}
where $\oslash$ denotes element-wise division and
$\mathbf R(\cdot)$ converts a unit quaternion into its corresponding
rotation matrix.






\begin{algorithm}[t]
\caption{Forward Prediction Path}
\label{alg:forward-path}
\begin{algorithmic}[1]
\Require Labeled initial state $Z_0$, requested timestamps ${t_i}$,
and object Gaussians $\mathcal{G}_0$
\Ensure Predicted object states ${Z_i}$ and, optionally, transformed
Gaussian proxies ${\mathcal{G}_i}$

\State Initialize $Z \gets Z_0$ and project its quaternion, scale,
and material channels onto their valid ranges

\For{each frame interval $[t_i,t_{i+1}]$}
\State Partition $[t_i,t_{i+1}]$ into RK4 substeps
\For{each RK4 substep}
\State Evaluate the structured vector field, advance $Z$ by one
RK4 step, and project it onto the valid state domain
\If{the horizontal-floor event is triggered}
\State Apply the analytic collision response and bounded residual
correction, then reproject $Z$
\EndIf
\EndFor
\State Store the projected state as $Z_{i+1}$
\EndFor

\If{Gaussian proxy prediction is required}
\For{each requested timestamp $t_i$}
\State Compute the pose and scale relative to $Z_0$ and construct
the corresponding affine map $A_i$
\State Apply $A_i$ to all Gaussians in $\mathcal{G}_0$ and store
the result as $\mathcal{G}_i$
\EndFor
\EndIf

\State \Return ${Z_i}$ and, if required, ${\mathcal{G}_i}$
\end{algorithmic}
\end{algorithm}

The Gaussian mean and covariance then evolve as
\begin{equation}
\mathbf g_i^t=\mathbf A_t(\mathbf g_i^0-\mathbf p_0)+\mathbf p_t,
\qquad
\boldsymbol\Sigma_i^t
=\mathbf A_t\boldsymbol\Sigma_i^0\mathbf A_t^\top.
\label{eq:gaussian-update}
\end{equation}
Color, opacity, and semantic attributes remain attached to the corresponding primitive, and background Gaussians remain unchanged.

This update has three useful properties. 
First, the initial and predicted rotation matrices together with the relative scale ratio give $\mathbf A_0=\mathbf I$ for a consistent initial state, since $\boldsymbol\rho_0=\mathbf 1$ and $\mathbf R_0\mathbf R_0^\top=\mathbf I$.
Second, centering at $\mathbf p_0$ separates object translation from rotation and scaling. Third, covariance congruence preserves positive semidefiniteness. The number of dynamics parameters and predicted trajectories is independent of the number of Gaussians, although applying Eq.~\eqref{eq:gaussian-update} and rendering still scale linearly with the number of primitives.

\subsection{Training and Inference}

The dynamics model is trained from labeled state sequences. Given $Z_0$ and $t_{1:T}$, it performs the same unrolled hybrid integration used at inference; ground-truth future states are loss targets and are never fed into the rollout. The training objective is
\begin{equation}
\mathcal L=\mathcal L_{\mathrm{state}}
+0.05\mathcal L_{\mathrm{pen}}
+0.01\mathcal L_{\mathrm{smooth}},
\label{eq:loss}
\end{equation}
with
\begin{equation}
\begin{aligned}
\mathcal L_{\mathrm{state}}={}&
\mathcal L_p+0.25\mathcal L_q+0.5\mathcal L_v
+0.2\mathcal L_\omega
+0.2\mathcal L_s+0.1\mathcal L_{\dot s} .
\end{aligned}
\end{equation}
$\mathcal L_p$, $\mathcal L_q$, $\mathcal L_v$, $\mathcal L_\omega$, $\mathcal L_s$, and $\mathcal L_{\dot{s}}$ constrain the object's position, orientation, linear velocity, angular velocity, scale, and scale rate, respectively; $\mathcal L_{\mathrm{pen}}$ penalizes floor penetration, while $\mathcal L_{\mathrm{smooth}}$ encourages temporal smoothness of the position and scale trajectories. \textit{Due to the page limitation, detailed formulations are provided in the Appendix.}
All non-orientation terms are component-wise mean squared errors, while
$\mathcal L_q=1-|\langle\hat{\mathbf q},\mathbf q\rangle|^2$ accounts for the quaternion sign ambiguity. The penetration term is
$\mathbb E[\operatorname{ReLU}(|s_y|-p_y)^2]$. The smoothness term penalizes $\mathcal L_{\mathrm{pen}}$ second temporal differences of position and, with relative weight $0.1$, scale. It is applied across the full sequence, including contact intervals. The reported objective contains no direct loss on $(m,e,\mu)$, Gaussian trajectories, rendered RGB, depth, masks, or Gaussian correspondences. At inference, the predicted states are passed through Eq.~\eqref{eq:gaussian-update} to obtain the dynamic Gaussian scene. The overall inference phase is shown in Algorithm \ref{alg:forward-path}.


\begin{table*}[t]
\centering
\small
\setlength{\tabcolsep}{4pt}
\begin{tabular*}{\textwidth}{@{\extracolsep{\fill}}llrrrrrr@{}}
\toprule
Split & Method & Traj $\downarrow$ & FDE $\downarrow$ & Vel $\downarrow$ & Quat $\downarrow$ & Scale $\downarrow$ & Plane Viol. $\downarrow$ \\
\midrule
\raisebox{-5.5ex}[0pt][0pt]{ID} 
 & Hold-$Z_0$ & 3.4104 & 3.9595 & 4.9891 & 0.3917 & \textbf{0.0256} & 0.062500 \\
 & Const-Vel-SE(3)          & 6.0296 & 8.5449 & 4.9891 & \textbf{0.0815} & 0.1386 & 0.047335 \\
 & Damped-Vel-SE(3)         & 4.7956 & 6.5425 & 4.1143 & 0.2590 & 0.1043 & 0.047339 \\
 & Gravity-Bounce-SE(3)     & 3.3238 & 4.1197 & 3.7467 & 0.2111 & 0.1386 & 0.001189 \\
 & Physics-Prior-SE(3)      & 3.2854 & 4.1143 & 3.7342 & 0.2330 & 0.2893 & 0.003909 \\
 & \textbf{NewtonGS (Ours) }   & \textbf{3.1669} & \textbf{3.8576} & \textbf{3.6469} & 0.1956 & 0.1626 & 0.002006 \\
\midrule
\raisebox{-5.5ex}[0pt][0pt]{OOD} 
 & Hold-$Z_0$ & 4.3073 & 4.8948 & 5.7827 & 0.3938 & \textbf{0.0256} & 0.062500 \\
 & Const-Vel-SE(3)          & 7.0243 & 9.8954 & 5.7827 & \textbf{0.0790} & 0.1387 & 0.047162 \\
 & Damped-Vel-SE(3)         & 5.4222 & 7.5384 & 4.5955 & 0.2799 & 0.1044 & 0.047164 \\
 & Gravity-Bounce-SE(3)     & 3.9078 & 4.8428 & 4.1008 & 0.2123 & 0.1387 & 0.001103 \\
 & Physics-Prior-SE(3)      & 3.8557 & 4.8298 & 4.0991 & 0.2386 & 0.2894 & 0.003620 \\
 & \textbf{NewtonGS (Ours) }   & \textbf{3.7377} & \textbf{4.5762} & \textbf{4.0059} & 0.2023 & 0.1608 & 0.001905 \\
\bottomrule
\end{tabular*}

\caption{State prediction results on the in-distribution (ID) and velocity-range-shift (OOD) splits of State-32. NewtonGS results are averaged over three independent runs, and its maximum across-run standard deviation is 0.0032. The analytic baselines are deterministic and seed-independent. Bold denotes the lowest mean error for each ranked metric within a split. Plane Viol.\ is reported as a diagnostic metric and is not ranked.}
\label{tab:state-prediction}

\end{table*}

\begin{table*}[t]
\centering
\small
\setlength{\tabcolsep}{4pt}
\begin{tabular*}{\textwidth}{@{\extracolsep{\fill}}lrrrrrr@{}}
\toprule
Method & Traj $\downarrow$ & FDE $\downarrow$ & Vel $\downarrow$ & Quat $\downarrow$ & Scale $\downarrow$ & Plane Viol. $\downarrow$ \\
\midrule
Hold-$Z_0$                & 2.6304 & 3.2268 & 4.8566 & 0.4132 & \textbf{0.0875} & 0.312500 \\
Gravity-Bounce-SE(3)      & 2.6747 & 3.3534 & 3.5683 & 0.1641 & 0.3666 & 0.010204 \\
Physics-Prior-SE(3)       & 2.6586 & 3.3399 & 3.5669 & 0.1720 & 0.3205 & 0.010204 \\
L4GM video-to-4DGS        & 2.9884 & 3.6727 & \textbf{2.9919} & 0.7676 & 0.3787 & 1.000000 \\
\textbf{NewtonGS (Ours) }                & \textbf{2.5751} & \textbf{3.2052} & 3.5116 & \textbf{0.1547} & 0.3129 & 0.010204 \\
\bottomrule
\end{tabular*}

\caption{Exploratory cross-representation comparison with L4GM on the fixed Same-32 subset. NewtonGS results are averaged over three independent runs, with a maximum standard deviation of 0.0132. The analytic baselines and L4GM evaluation are deterministic. Bold denotes the lowest mean error in each ranked metric, while Plane Viol.\ is diagnostic and unranked.}
\label{tab:main-video-to-4dgs}

\end{table*}

\begin{table*}[t]
\centering
\small
\setlength{\tabcolsep}{8pt}
\resizebox{\textwidth}{!}{%
\begin{tabular}{llrrrr}
\toprule
Method & Evaluated output & $N$ & ${\rm PIS}_p\uparrow$ & ${\rm BC}_p\uparrow$ & ${\rm MS}_p\uparrow$ \\
\midrule
PhyT2V~\citep{xue2025phyt2v} & released single-view video & 12 & \meanstd{0.6914}{0.0770} & \meanstd{0.8112}{0.1967} & \meanstd{0.6862}{0.2120} \\
Sora~\citep{brooks2024sora} & released single-view video & 12 & \meanstd{0.6185}{0.1003} & \meanstd{0.5967}{0.2175} & \meanstd{0.5035}{0.2270} \\
Veo~3~\citep{google2025veo3} & released single-view video & 12 & \meanstd{0.6458}{0.0891} & \meanstd{0.7911}{0.2117} & \meanstd{0.6204}{0.2170} \\
Wan~\citep{wanteam2025wan} & seeded single-view video & 12 & \meanstd{0.7109}{0.0912} & \meanstd{0.8727}{0.1852} & \meanstd{0.7678}{0.2350} \\
\textbf{NewtonGS (Ours) }& rendered 4DGS view & 12 & \meanstd{0.7322}{0.0426} & \meanstd{0.9231}{0.0720} & \meanstd{0.8679}{0.0953} \\
\bottomrule
\end{tabular}
}

\caption{Common-12 proxy comparison with video-generation methods across 12 coarsely aligned motion categories. Values are mean $\pm$ sample s.d.\ over one video per category. NewtonGS and Wan use fixed-seed generations, while PhyT2V, Sora, and Veo use their released videos.}
\label{tab:main-video-proxy-common12}

\end{table*}

\section{Experiments}

In this section, we evaluate the proposed NewtonGS.


\subsection{Datasets}

Procedurally generated benchmarks have been widely used to provide controlled physical interactions, exact state supervision, and systematic generalization splits \cite{bakhtin2019phyre, bear2021physion}. Following this evaluation paradigm, we use two synthetic datasets to evaluate NewtonGS. State-32 is the primary benchmark for object-state rollout, whereas Gaussian-32 evaluates the state-to-Gaussian transformation and cross-representation interfaces. \textit{Complete generation procedures and details are provided in the Appendix.}

\noindent \textbf{State-32.} State-32 contains 1,048,576 training sequences and 131,072 sequences in each of the in-distribution (ID) and out-of-distribution (OOD) evaluation splits. Each sequence consists of 64 object states sampled at 24Hz and belongs to one of 32 procedurally defined motion families, whose labels are not provided to the model. The ID split is sampled from the same parameter ranges as the training set. The OOD split preserves the motion families and all non-velocity parameter ranges, but increases the maximum horizontal, vertical, and angular velocities to 1.6 times their corresponding training limits. State-32 contains state trajectories only and provides no appearance, Gaussian, or rendering supervision.

\noindent \textbf{Gaussian-32.} Gaussian-32 contains 320 appearance-conditioned synthetic sequences obtained by combining 10 object--scene templates with the same 32 motion families used in State-32. Each sequence provides Gaussian primitives, camera parameters, and rendering information. We additionally define a fixed \emph{Same-32} subset containing one sequence from each motion family for comparisons with methods that operate on videos or independently reconstructed 4D Gaussian representations. The Same-32 subset is separate from the ID and OOD splits of State-32 and is not used to train NewtonGS.

\subsection{Implementation Details}
\noindent \textbf{Training setup.}
The State-32 experiments use labeled initial states $Z_0$, whereas Gaussian grouping and state lifting are used only for Gaussian-32. We train models for 100 epochs with three independent runs using seeds 7301, 7302, and 7303, respectively. The model has 81,429 trainable parameters and is trained solely with the state-space objective in Eq.~\eqref{eq:loss}, without Gaussian-level or rendering supervision. Checkpoints are selected on a held-out ID validation set and evaluated on the ID and OOD test sets. \textit{Additional training details are provided in the Appendix.}

\noindent \textbf{Network and rollout.}
The continuous and contact residuals use MLPs with dimensions $22\!\rightarrow\!256\!\rightarrow\!256\!\rightarrow\!9$ and $22\!\rightarrow\!256\!\rightarrow\!6$, respectively. Both use SiLU activations, Xavier-initialized hidden layers~\citep{glorot2010xavier}, and zero-initialized output layers. Rollouts use RK4 with a maximum step of $1/30$\,s, resulting in two $1/48$\,s substeps per frame at 24\,Hz. After each substep, we normalize the quaternion, project bounded state variables to their valid ranges, and apply the horizontal-floor contact test using $|s_y|$ as the vertical support-radius proxy.

\noindent \textbf{Gaussian grouping and state lifting.}
We project each Gaussian mean into the binary object masks across all cameras, discard projections with nonpositive depth, and assign Gaussians whose average mask membership is at least $0.5$. The deterministic state lifter estimates the object center, orientation, and aggregate scale using opacity- and membership-weighted PCA~\citep{jolliffe2002pca}. Motion rates are estimated by finite differences when multiple frames are available and initialized to zero otherwise. Material parameters are obtained from metadata or fixed defaults.

\subsection{Evaluation Metrics}

We compute the state-prediction metrics over all evaluated sequences and frames. Trajectory RMSE (Traj) is the root mean square of the Euclidean position error, while final displacement error (FDE) is the mean Euclidean position error at the final frame. Velocity RMSE (Vel) is defined analogously using the predicted and ground-truth velocities, and Scale RMSE is computed over the individual scale components. Quaternion error (Quat) is the mean bounded, sign-invariant discrepancy $1-|\langle\hat{\mathbf q},\mathbf q\rangle|^2$, and it lies in $[0,1]$. Position, velocity, and scale errors are reported in synthetic generator units rather than physical units such as meters. Plane Viol.\ denotes the fraction of predicted states satisfying $|s_y|-p_y>10^{-6}$ under the horizontal-floor proxy. Because it measures consistency with the prescribed floor constraint rather than agreement with the target trajectory, we report it without ranking or bolding methods. $\mathrm{PIS}_p$ measures physical consistency, $\mathrm{BC}_p$ measures background stability, and $\mathrm{MS}_p$ measures the continuity and smoothness of motion.

\subsection{Baselines and Comparison Protocol}

All state-space methods are initialized with the same labeled state $Z_0$, evaluated at the same timestamps, and they do not receive future states or the hidden motion-family label. Hold-$Z_0$ repeats the initial state at every timestamp. Const-Vel-SE(3) extrapolates translation, orientation, and scale using their initial rates, whereas Damped-Vel-SE(3) applies exponential decay to these rates. Gravity-Bounce-SE(3) additionally incorporates gravity, semi-implicit Euler integration, and the analytic horizontal-floor response used by NewtonGS, but does not include a learned contact residual. Physics-Prior-SE(3) further introduces fixed linear and angular damping together with a scale-restoring acceleration.
L4GM \cite{ren2024l4gm} reconstructs a time-varying 4DGS with its own Gaussian topology from foreground video, rather than receiving labeled $Z_0$ or predicting State-32-aligned trajectories. Public-video methods generate single-view RGB videos from method-specific conditions; because their inputs and outputs differ from NewtonGS, these methods are evaluated separately as cross-representation interface checks.

\begin{figure*}[htbp]
\centering
\includegraphics[width=1\textwidth]{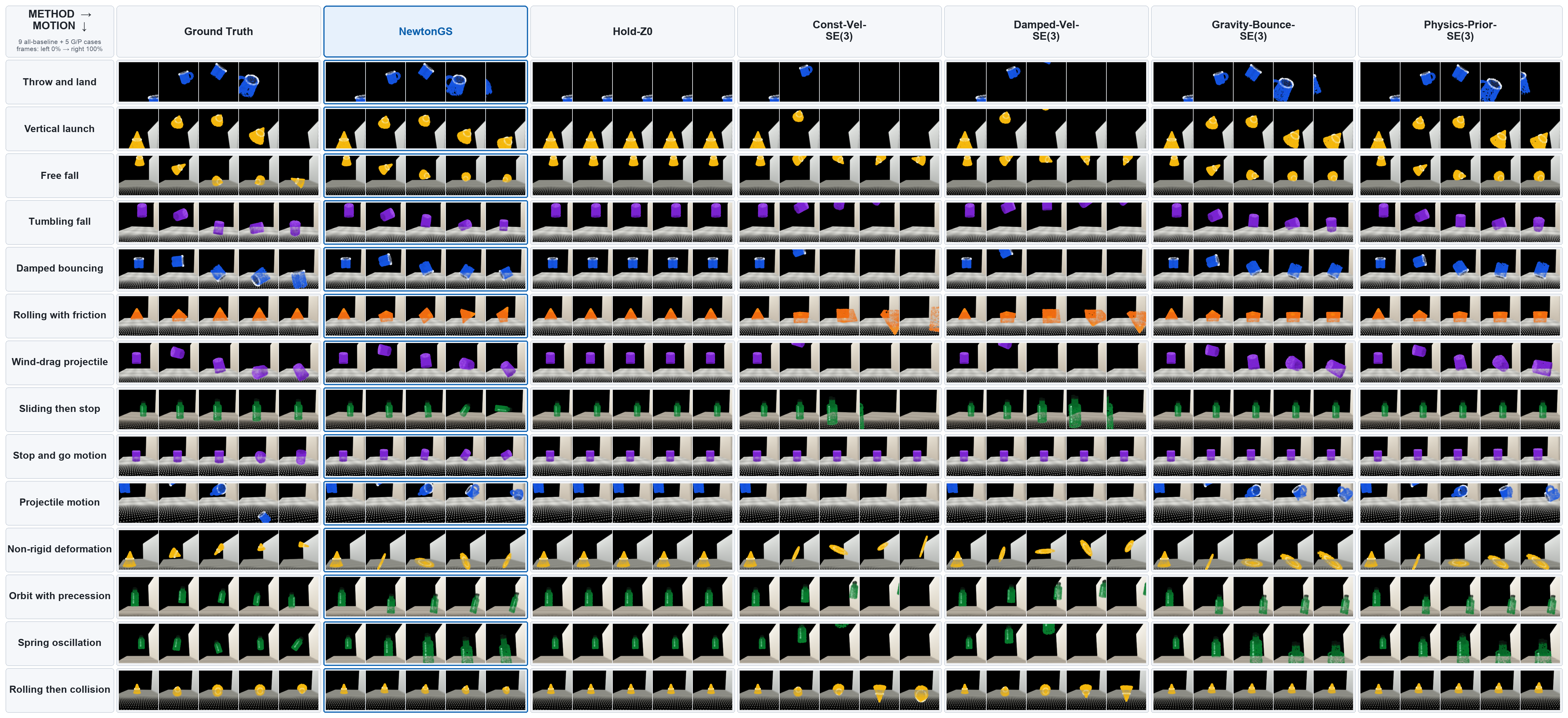}

\caption{Visualization of different methods. Blue identifies the proposed NewtonGS.}
\label{fig:table2-rendered-samples}
\end{figure*}

\subsection{Quantitative Results and Analysis}

Table~\ref{tab:state-prediction} evaluates 131,072 sequences on each of the full ID and OOD splits of State-32. NewtonGS achieves the lowest trajectory RMSE, final displacement error, and velocity RMSE on both splits, ranking first in all six core translational-dynamics comparisons. Relative to the best-performing baseline in each column, NewtonGS reduces the trajectory, endpoint, and velocity errors by $3.61\%$, $2.57\%$, and $2.34\%$ on ID, and by $3.06\%$, $5.25\%$, and $2.27\%$ on OOD, respectively. NewtonGS maintains its lead when moving from ID to OOD with an expanded velocity range, indicating that its advantage is not confined to the training parameter range. Moreover, the maximum standard deviation across its three training runs is only $0.0032$, substantially smaller than the performance gaps to the competing methods, demonstrating good run-to-run stability.
NewtonGS does not achieve the lowest quaternion or scale error because Const-Vel-SE(3) directly extrapolates the initial angular velocity, whereas Hold-$Z_0$ keeps the initial scale unchanged; these strong priors specifically favor their corresponding metrics. Plane Viol. measures only compatibility with a simplified horizontal-floor constraint rather than prediction accuracy and is therefore excluded from the method ranking. Overall, the results show that the structured continuous dynamics and learned residuals of NewtonGS consistently improve trajectory, endpoint, and velocity prediction under both ID and OOD conditions, constituting its main advantage over fixed analytic dynamics.

\subsection{Comparison with Video-to-4DGS Methods}
Table~\ref{tab:main-video-to-4dgs} shows that NewtonGS achieves the lowest trajectory RMSE, final displacement error, and quaternion error. Relative to the best-performing baseline in each corresponding column, NewtonGS reduces three errors by $2.10\%$, $0.67\%$, and $5.73\%$, respectively. Compared with Physics-Prior-SE(3), it reduces the trajectory, endpoint, velocity, quaternion, and scale errors by $3.14\%$, $4.03\%$, $1.55\%$, $10.06\%$, and $2.37\%$, respectively, demonstrating that learned residuals consistently improve individual state components beyond the structured physics prior.
Although L4GM achieves the lowest velocity error and Hold-\(Z_0\) achieves the lowest scale error, NewtonGS provides a more balanced overall performance across position, orientation, velocity, and scale.

\subsection{Comparison with Video Generation Methods}
Table~\ref{tab:main-video-proxy-common12} aligns the number of evaluation categories, sample count, and proxy evaluator under the common-12 protocol. NewtonGS achieves the highest mean on all metrics while exhibiting the lowest cross-video standard deviations, indicating that its advantages generalize consistently across motion categories. Compared with Wan, the strongest single-view video method in the table, NewtonGS improves \({\rm PIS}_p\), \({\rm BC}_p\), and \({\rm MS}_p\) by $3.00\%$, $5.78\%$, and $13.03\%$, respectively. The corresponding standard deviations are reduced by $53.29\%$, $61.12\%$, and $59.45\%$, demonstrating more consistent performance across motion categories.
These results indicate that the structured dynamics and time-varying 4DGS representation of NewtonGS produce renderings with greater motion plausibility, background consistency, and motion stability.


Figure~\ref{fig:table2-rendered-samples} shows 14 motion cases. Across the five sampled frames, NewtonGS accurately preserves the direction, curvature, and temporal evolution of the ground-truth motion. For ballistic motion, it captures acceleration and landing; for bouncing and rolling motion, it more accurately predicts changes in contact states; and for oscillatory and orbital motion, it recovers curved or reversing trajectories. In contrast, the analytic baselines often remain static, drift linearly, over-damp the motion, or predict incorrect contact timing. These results demonstrate that, by combining structured dynamics with learned residual corrections, NewtonGS can model diverse motion patterns that fixed analytic rules fail to capture.

\section{Conclusion}
We present NewtonGS, an object-level dynamics framework that assigns explicit states to Gaussian objects, evolves them through structured continuous dynamics, learned residuals, and discrete contact events, and converts the predicted pose and scale into shared transformations of the associated Gaussian primitives.  On the procedurally generated State-32 dataset, NewtonGS achieves lower trajectory, final displacement, and velocity errors than five analytic baselines on both the ID and velocity-range-shift OOD splits. Results on Gaussian-32 further illustrate state-driven Gaussian object animation. Future work will explore automatic state estimation from videos, richer contact and multi-object interactions, articulated and topology-changing motion, and evaluation on real multi-view scenes and learned baselines.
\clearpage

\nocite{ye2025gsplat}
\bibliographystyle{iclr2026_conference}
\bibliography{aaai2027}

\clearpage
\appendix
\section*{Appendix}

This appendix is organized as follows. 
\begin{itemize}
    \item \textbf{Section~A} documents the model
objective, state conventions, Gaussian transformation, initialization,
optimization, evaluation, and inference.
\item \textbf{Section~B} specifies the Gaussian-32
evaluation set, State-32 procedural corpus, Same-32 fixed subset, and
Common-12 manifest.
\item  \textbf{Section~C} specifies the L4GM/Same-32
cross-representation evaluation, the Common-12 video proxy comparison,
and analytic baselines.
\item \textbf{Section~D} reports the residual and branch ablations.
\item \textbf{Section~E} presents motion-wise results, parameter-response tests, and the
PIS-3D diagnostic.
\item \textbf{Section~F} presents qualitative comparisons and
Gaussian-scene rollouts.
\item \textbf{Section~G} defines the method's applicability,
failure modes, and observation-to-state gap.
\end{itemize}

\section{Implementation Details}
This section specifies the exact implementation used in all reported experiments. We describe the training objective, state representation, Gaussian-to-state lifting and state-to-Gaussian transformation, parameter initialization, optimization, evaluation metrics, and inference procedure.

\subsection{Training Loss}

For any vector-valued state field $x$ with $d_x$ components, the implementation defines
\begin{equation}
\operatorname{MSE}_x=
\frac{1}{BTd_x}\sum_{b=1}^{B}\sum_{t=1}^{T}
\|\widehat{\mathbf x}_{b,t}-\mathbf x_{b,t}\|_2^2.
\end{equation}
Here $Z$ is the ground-truth state tensor, $\widehat Z$ is its prediction, and $\widehat Z,Z\in\mathbb R^{B\times T\times22}$.
$B$ is batch size, $T$ is the number of sampled timestamps, and 22 is the
number of channels in one state. The indices $b\in\{1,\ldots,B\}$ and
$t\in\{1,\ldots,T\}$ identify a batch element and a timestamp.
$\mathbf x_{b,t}\in\mathbb R^{d_x}$ and
$\widehat{\mathbf x}_{b,t}\in\mathbb R^{d_x}$ are the target and predicted
values of field $x$ at that entry; $d_x$ is the number of scalar components in
that field. The two summation signs aggregate all batch and time entries,
$\|\cdot\|_2^2$ is squared Euclidean distance over the $d_x$ components, and
the denominator $BTd_x$ makes $\operatorname{MSE}_x$ a per-component mean
squared error.

The sign-invariant quaternion term is reduced over batch and time,
\begin{equation}
\mathcal L_q=\frac1{BT}\sum_{b,t}
\left(1-\left|\left\langle
\frac{\widehat{\mathbf q}_{b,t}}{\|\widehat{\mathbf q}_{b,t}\|_2},
\frac{\mathbf q_{b,t}}{\|\mathbf q_{b,t}\|_2}
\right\rangle\right|^2\right),
\end{equation}
where $\mathcal L_q$ is the orientation loss and
$\widehat{\mathbf q}_{b,t}$ and $\mathbf q_{b,t}$ are the predicted and target
quaternions. Each fraction normalizes a quaternion to unit length,
$\langle\cdot,\cdot\rangle$ is the Euclidean inner product in
$\mathbb R^4$, $|\cdot|$ removes the physically irrelevant sign ambiguity
between $\mathbf q$ and $-\mathbf q$, and the outer square converts alignment
to a bounded discrepancy. The shorthand $\sum_{b,t}$ means summation over all
$b=1,\ldots,B$ and $t=1,\ldots,T$. Both normalizations use a $10^{-8}$
denominator floor, and the absolute inner product is clamped to $[0,1]$ before
squaring.

The executed state loss is
\begin{equation}
\begin{split}
\mathcal L_{\mathrm{state}}={}&
1.0\operatorname{MSE}_{p}
+0.25\mathcal L_q
+0.5\operatorname{MSE}_{v}\\
&+0.2\operatorname{MSE}_{\omega}
+0.2\operatorname{MSE}_{s}
+0.1\operatorname{MSE}_{u}\\
&+0.0\operatorname{MSE}_{m,e,\mu}.
\end{split}
\label{eq:appendix-state-loss}
\end{equation}
In Eq.~\ref{eq:appendix-state-loss}, $\mathcal L_{\mathrm{state}}$ is the
weighted state-reconstruction loss; $\mathbf p$, $\mathbf v$,
$\boldsymbol\omega$, $\mathbf s$, and $\mathbf u$ denote position,
linear velocity, angular velocity, scale, and scale rate. The terms
$\operatorname{MSE}_{p}$, $\operatorname{MSE}_{v}$,
$\operatorname{MSE}_{\omega}$, $\operatorname{MSE}_{s}$, and
$\operatorname{MSE}_{u}$ use the preceding MSE definition on those
fields. $\operatorname{MSE}_{m,e,\mu}$ applies it to the concatenated material
tuple $(m,e,\mu)$. The numerical coefficients
$1.0,0.25,0.5,0.2,0.2,0.1,$ and $0.0$ are fixed loss weights. In particular,
the zero coefficient means that material channels are carried in the state but
receive no direct reconstruction penalty.

For a floor at $y_f=0$, the two regularizers are
\begin{equation}
\begin{split}
\mathcal L_{\mathrm{pen}}&=\frac1{BT}\sum_{b,t}
\operatorname{ReLU}(y_f+|\widehat s_{y,b,t}|-\widehat p_{y,b,t})^2,\\
\mathcal L_{\mathrm{smooth}}&=
\operatorname{mean}\!\left[(\Delta_t^2\widehat{\mathbf p})^2\right]
+0.1\operatorname{mean}\!\left[(\Delta_t^2\widehat{\mathbf s})^2\right].
\end{split}
\end{equation}
Here $\mathcal L_{\mathrm{pen}}$ is the floor-penetration penalty and
$\mathcal L_{\mathrm{smooth}}$ is the temporal smoothness penalty.
$y_f$ is the vertical coordinate of the floor; $\widehat p_{y,b,t}$ is the
predicted vertical position and $\widehat s_{y,b,t}$ is the predicted vertical
scale component. The absolute value $|\widehat s_{y,b,t}|$ supplies a
nonnegative support extent. $\operatorname{ReLU}(z)=\max(0,z)$ retains only
positive penetration depth, and its square penalizes larger violations more
strongly. $\widehat{\mathbf p}$ and $\widehat{\mathbf s}$ are the full
predicted position and scale sequences. The temporal second-difference
operator is
$\Delta_t^2x_t=x_{t+1}-2x_t+x_{t-1}$; it measures discrete curvature rather
than velocity. Squaring a vector inside \texttt{mean} is component-wise, and
each mean includes all valid batch, interior-time, and vector-component
entries. The coefficient $0.1$ gives scale smoothness one tenth of the weight
of position smoothness.

The complete optimized objective is
\begin{equation}
\mathcal L=\mathcal L_{\mathrm{state}}
+0.05\mathcal L_{\mathrm{pen}}
+0.01\mathcal L_{\mathrm{smooth}}.
\label{eq:appendix-total-loss}
\end{equation}
In Eq.~\ref{eq:appendix-total-loss}, $\mathcal L$ is the scalar objective
minimized by the optimizer. The three addends are the state, penetration, and
smoothness losses defined above, while $1$, $0.05$, and $0.01$ are their fixed
global weights.
All 64 timestamps, including the supplied initial state and contact intervals, participate.

\subsection{State Conventions}

For the state defined in the main paper, the exact zero-based implementation
slices are position 0-2, quaternion 3-6, velocity 7-9, angular velocity
10-12, scale 13-15, scale rate 16-18, and material variables 19-21.
Equivalently, one state contains
$(\mathbf p,\mathbf q,\mathbf v,\boldsymbol\omega,\mathbf s,\mathbf u,m,e,\mu)$:
$\mathbf p,\mathbf v,\boldsymbol\omega,\mathbf s,\mathbf u\in\mathbb R^3$
are position, linear velocity, angular velocity, scale, and scale rate;
$\mathbf q\in\mathbb R^4$ is orientation; and $m,e,\mu\in\mathbb R$ are
mass, restitution, and tangential attenuation. Their dimensions sum to 22.

The loader accepts a tensor of shape $(T,C)$ or $(B,T,C)$, where $B$ is batch
size, $T$ is the number of timestamps, and $C$ is the number of supplied
channels. It adds a batch dimension when necessary, retains the first 22
channels, converts the result to contiguous float32, and performs no
dataset-level centering, standardization, or whitening. A state projection is
applied initially, at the intermediate RK4 stages~\citep{hairer1993ode}, after
every RK4 step, and after contact response:
\begin{equation}
\begin{aligned}
\|\mathbf q\|_2&=1, & s_j&\geq10^{-4},\\
m&\geq10^{-4}, & e,\mu&\in[0,1].
\end{aligned}
\end{equation}
Here $\mathbf q\in\mathbb R^4$ is the orientation quaternion and
$\|\mathbf q\|_2$ is its Euclidean norm. The symbol $s_j$ denotes component
$j\in\{x,y,z\}$ of the three-dimensional scale vector $\mathbf s$.
The scalars $m$, $e$, and $\mu$ denote mass, normal restitution, and
tangential attenuation, respectively. The constant $10^{-4}$ is the numerical
lower bound used to prevent zero scale or mass, and $[0,1]$ is the closed
admissible interval for $e$ and $\mu$.

Quaternions are scalar-first, $\mathbf q=[q_w,q_x,q_y,q_z]$, where $q_w$ is
the real component and $(q_x,q_y,q_z)$ are the three imaginary components;
multiplication is the Hamilton product. In particular, the executed kinematic
update is
\begin{equation}
\dot{\mathbf q}=\tfrac12\mathbf q\otimes[0,\boldsymbol\omega].
\label{eq:appendix-quaternion-order}
\end{equation}
In Eq.~\ref{eq:appendix-quaternion-order}, $\dot{\mathbf q}=d\mathbf q/dt$
is the time derivative of orientation, $t$ denotes continuous time,
$\boldsymbol\omega=(\omega_x,\omega_y,\omega_z)\in\mathbb R^3$ is angular
velocity, $[0,\boldsymbol\omega]$ is the pure quaternion with zero real part,
$\otimes$ denotes the Hamilton product, and the factor $1/2$ is the standard
quaternion kinematic coefficient under this multiplication order.

\subsection{Gaussian Grouping and State Lifting}

This deterministic interface draws on the explicit scene representation of
3D Gaussian Splatting~\citep{kerbl2023gs}, the standard covariance-eigenvector
interpretation of principal component analysis~\citep{jolliffe2002pca}, and
the unit-quaternion convention of Shoemake~\citep{shoemake1985quaternion}.
It is used only for the Gaussian experiments; the state-only training
benchmark starts directly from labeled initial states.

\subsection{Object-Level Gaussian Transformation}

For the state-to-Gaussian update defined in the main paper, the implementation constructs its affine matrix as
\begin{equation}
\begin{aligned}
\mathbf R_{\mathrm{rel},t} &= \mathbf R(\mathbf q_t\otimes\mathbf q_0^*),\\
\boldsymbol\rho_t &=
(\mathbf s_t\oslash\max(\mathbf s_0,10^{-6}))_{\geq10^{-6}},\\
\mathbf A_t^{\mathrm{impl}} &=
\mathbf R_{\mathrm{rel},t}\operatorname{diag}(\boldsymbol\rho_t).
\end{aligned}
\label{eq:appendix-gaussian-update}
\end{equation}
In Eq.~\ref{eq:appendix-gaussian-update}, subscript $0$ denotes the initial
state and subscript $t$ the state at time $t$. The function $\mathbf R(\cdot)$
maps a unit quaternion to its $3\times3$ rotation matrix;
$\mathbf q_0^*$ is the conjugate (and, for a unit quaternion, inverse) of the
initial orientation; and $\mathbf q_t\otimes\mathbf q_0^*$ is the relative
orientation. Thus $\mathbf R_{\mathrm{rel},t}$ is the corresponding relative
rotation matrix. The vector $\boldsymbol\rho_t\in\mathbb R^3$ contains the
component-wise scale ratios, $\oslash$ denotes component-wise division, and
$\max(\mathbf s_0,10^{-6})$ applies the lower bound separately to all three
entries. The subscript ${\geq10^{-6}}$ means that the resulting ratios are
again clamped component-wise from below. The operator
$\operatorname{diag}(\boldsymbol\rho_t)$ places the three ratios on the
diagonal of a $3\times3$ matrix. Finally,
$\mathbf A_t^{\mathrm{impl}}\in\mathbb R^{3\times3}$ is the implemented
linear transform, and the superscript ``impl'' distinguishes it from an
idealized local-frame transform.
Equation~\ref{eq:appendix-gaussian-update} is the column-vector equivalent of the row-vector implementation: centered means are scaled component-wise in their stored coordinate components and then rotated by the relative orientation. For nonidentity $\mathbf q_0$, this is not equivalent to anisotropic scaling in a continuously rotating local object frame. If no initial state is supplied, the fallback uses the identity quaternion, unit scale, and mean Gaussian center. The sequence converter applies the mean and covariance updates from the main paper using $\mathbf A_t^{\mathrm{impl}}$ and emits covariance when it is present.

\subsection{Parameter Initialization}

The trainable continuous-dynamics scalars are initialized as
$c_v=c_\omega=0.05$, $k_s=0.25$, $c_s=0.08$, and $\alpha=0.01$.
Here $c_v$ and $c_\omega$ control linear and angular damping, $k_s$ controls
scale restoration, $c_s$ controls scale-rate damping, and $\alpha$ scales the
continuous learned residual. The implementation applies absolute values to
$c_v,c_\omega,k_s,$ and $c_s$ so that their effective coefficients are
nonnegative, whereas $\alpha$ retains its learned sign.

For both residual networks, hidden-layer weights use Xavier uniform
initialization and hidden-layer biases are initialized to zero. Output-layer
weights and biases are also initialized to zero, making both residual
functions exactly zero at initialization while still allowing gradients to
update their output layers. The contact-residual scale
$\alpha_{\mathrm{imp}}$ is initialized to $0.01$ and, like $\alpha$, retains
its sign rather than being passed through an absolute value.

\subsection{Optimization}

\begin{table*}[!ht]
\centering
\small
\begin{tabular}{L{0.24\textwidth}L{0.68\textwidth}}
\toprule
Setting & Executed value \\
\midrule
Optimizer & AdamW, learning rate $10^{-4}$, default $\beta=(0.9,0.999)$, $\epsilon=10^{-8}$, weight decay $0.01$, no AMSGrad \\
Schedule & Cosine annealing with $T_{\max}=100$ and $\eta_{\min}=10^{-6}$; scheduler state restored on resume \\
Updates & 16 optimizer updates per epoch \\
Randomness & For each run seed $s$ listed in the main paper, Python, NumPy, and CPU PyTorch use $s$, and the epoch permutation uses $s+\text{epoch}$ \\
Checkpointing & Epoch 1 and every 5 epochs; best checkpoint selected by the complete \texttt{val\_id} objective; epoch 100 selected in all three runs \\
Numerics & Float32; no automatic mixed precision; no gradient clipping \\
\bottomrule
\end{tabular}
\caption{Optimization settings shared by all three independently initialized runs.}
\label{tab:appendix-train-config}
\end{table*}
In Table~\ref{tab:appendix-train-config}, $\beta=(\beta_1,\beta_2)$ contains
AdamW's first- and second-moment decay coefficients, $\epsilon$ is its
denominator-stabilization constant, and weight decay is the decoupled
$\ell_2$ regularization coefficient. For the cosine scheduler, $T_{\max}$ is
the 100-epoch cycle length and $\eta_{\min}$ is the terminal learning rate.
In the randomness row, $s$ is the run seed, and $s+\text{epoch}$ makes the
epoch permutation deterministic and reproducible.

\subsection{Evaluation Execution}

Prediction uses batches of at most 65,534 trajectories and GPU data
parallelism. The larger evaluator output also contains a ground-truth oracle,
which is omitted from method comparisons.

For completeness, let $\mathbf e^p_{b,t}=\widehat{\mathbf p}_{b,t}-\mathbf
p_{b,t}$ and $\mathbf e^v_{b,t}=\widehat{\mathbf v}_{b,t}-\mathbf v_{b,t}$. The
executed aggregate metrics are
\begin{equation}
\begin{aligned}
\mathrm{Traj}&=\sqrt{\frac1{BT}\sum_{b,t}\|\mathbf e^p_{b,t}\|_2^2},\\
\mathrm{FDE}&=\frac1B\sum_b\|\mathbf e^p_{b,T}\|_2,\\
\mathrm{Vel}&=\sqrt{\frac1{BT}\sum_{b,t}\|\mathbf e^v_{b,t}\|_2^2},\\
\mathrm{Scale}&=\sqrt{\frac1{3BT}\sum_{b,t,j}
(\widehat s_{b,t,j}-s_{b,t,j})^2},\\
\mathrm{AngVel}&=\sqrt{\frac1{3BT}\sum_{b,t,j}
(\widehat\omega_{b,t,j}-\omega_{b,t,j})^2},\\
\mathrm{Quat}&=\frac1{BT}\sum_{b,t}
\left(1-|\langle\widehat{\mathbf q}_{b,t},\mathbf q_{b,t}\rangle|^2\right).
\end{aligned}
\label{eq:appendix-state-metrics}
\end{equation}
In Eq.~\ref{eq:appendix-state-metrics}, $B$ is the number of evaluated
trajectories, $T$ is the number of timestamps per trajectory,
$b\in\{1,\ldots,B\}$ indexes trajectories, $t\in\{1,\ldots,T\}$ indexes
timestamps, and $j\in\{x,y,z\}$ indexes the three Cartesian components.
A hat marks a prediction and an unhatted symbol its target.
$\mathbf e^p_{b,t}$ and $\mathbf e^v_{b,t}$ are the three-dimensional position
and velocity error vectors defined immediately above the equation.
$\|\cdot\|_2$ is Euclidean norm, the summations average the indicated samples,
and the square roots convert mean squared errors back to the units of the
underlying fields. $\mathrm{Traj}$ is position RMSE over the full trajectory;
$\mathrm{FDE}$ is mean final displacement error at terminal index $T$;
$\mathrm{Vel}$ is velocity RMSE; $\mathrm{Scale}$ and $\mathrm{AngVel}$ are
component-wise RMSEs for scale and angular velocity. The factor 3 in their
denominators accounts for the three vector components.

$\mathrm{Quat}$ is mean sign-invariant quaternion discrepancy.
$\widehat{\mathbf q}_{b,t}$ and $\mathbf q_{b,t}$ are normalized predicted and
target quaternions, $\langle\cdot,\cdot\rangle$ is their four-dimensional
inner product, and the absolute value identifies the equivalent signs
$\mathbf q$ and $-\mathbf q$. The legacy evaluator key
\texttt{quaternion\_geodesic} therefore denotes this bounded discrepancy, not
a geodesic angle in radians. For the contact diagnostics,
$d_{b,t}=\operatorname{ReLU}(|\widehat s_{y,b,t}|-\widehat p_{y,b,t})$ is
predicted penetration depth relative to the zero-height floor:
$\widehat s_{y,b,t}$ is predicted vertical scale,
$\widehat p_{y,b,t}$ is predicted vertical position, and ReLU retains only
positive violations. Penetration MSE is
$\frac1{BT}\sum_{b,t}d_{b,t}^2$. Plane Viol.\ is
$\frac1{BT}\sum_{b,t}\mathbb1[d_{b,t}>10^{-6}]$, where $\mathbb1$ is the
indicator function and $10^{-6}$ is the numerical violation threshold.

For class $c$, the code computes the class diagnostic error $E_c$ described
below and
\begin{equation}
\mathrm{PIS\mbox{-}3D}=\sum_c\frac{n_c}{B}\exp(-E_c).
\end{equation}
Here $\mathrm{PIS\mbox{-}3D}$ is the aggregate diagnostic score, $c$ indexes
the motion classes present in the evaluated split, $E_c\geq0$ is the
class-specific diagnostic error defined in
Table~\ref{tab:appendix-pis3d-formulas}, and $n_c$ is the number of evaluated
trajectories in class $c$. The total number of trajectories is
$B=\sum_c n_c$, so $n_c/B$ is the empirical class weight. The exponential
$\exp(-E_c)$ maps zero error to one and monotonically decreases toward zero as
the error increases. The outer sum forms a class-frequency-weighted mean.
This quantity is not a target-state distance and is not included in the ranked
main metrics.

For each motion class $c$, PIS-3D computes a class-specific nonnegative error $E_c$ and returns $\exp(-E_c)$ with coefficient one. The implemented diagnostics are normalized speed variance for uniform motion; horizontal-velocity variance and vertical-acceleration error for projectile, gravity, or acceleration labels; radius and angular-speed variance for circular or orbital labels; angular-speed variance for rotation; penetration plus observed-restitution error for bouncing or collision; acceleration error for slope motion using a fixed 22-degree slope; scale-rate norm variance for size change; normalized volume variance for deformation; and acceleration norm for unmatched labels. Split-level PIS-3D is the sample-count-weighted mean of per-class scores. Since several terms assess invariance rather than target displacement, a stationary sequence can score highly; PIS-3D is therefore treated as a diagnostic.

For an exact specification, let
$\mathcal V(x)=B^{-1}\sum_b\operatorname{Var}_t(x_{b,t})$, where the executed
\texttt{torch.var} uses its default sample-variance correction; let
$\mathcal M(x)$ denote the mean over all supplied entries; and let
$\mathbf a_t=(\mathbf v_{t+1}-\mathbf v_{t-1})/(2h)$.
Table~\ref{tab:appendix-pis3d-formulas} gives the class error selected by
substring matching.
\begin{table*}[!ht]
\centering
\small
\begin{tabular}{L{0.18\textwidth}L{0.74\textwidth}}
\toprule
Motion-name trigger & Executed $E_c$ \\
\midrule
\texttt{uniform} &
$\mathcal V(\|\mathbf v\|)/[\mathcal M(\|\mathbf v\|)^2+10^{-6}]$ \\
\texttt{projectile}, \texttt{gravity}, or \texttt{acceleration} &
$\mathcal V(v_x)+\mathcal V(v_z)+\mathcal M(|a_y+9.81|)$ \\
\texttt{circular} or \texttt{orbital} &
$\mathcal V(\|\mathbf p-\mathcal M_t(\mathbf p)\|)+\mathcal V(\|\boldsymbol\omega\|)$ \\
\texttt{rotation} & $\mathcal V(\|\boldsymbol\omega\|)$ \\
\texttt{bouncing} or \texttt{collision} &
$\mathcal M[\operatorname{ReLU}(|s_y|-p_y)]$, plus
$|\,\mathcal M_{\mathrm{sign\ flip}}[v_y^+/(-v_y^-)]-\mathcal M(e)\,|$ if at
least one negative-to-positive $v_y$ transition exists \\
\texttt{slope} &
$\mathcal M\{|a_x-[9.81\sin22^\circ-\mathcal M(\mu)9.81\cos22^\circ]\cos22^\circ|\}$ \\
\texttt{size} & $\mathcal V(\|\mathbf u\|)$ \\
\texttt{deformation} &
$\mathcal V(s_xs_ys_z)/[\mathcal M(s_xs_ys_z)^2+10^{-6}]$ \\
otherwise & $\mathcal M(\|\mathbf a\|)$ \\
\bottomrule
\end{tabular}
\caption{Exact PIS-3D class-error dispatch. The first matching branch is used;
the diagnostic is computed from predictions only and does not compare with the
target state.}
\label{tab:appendix-pis3d-formulas}
\end{table*}

Here $x_{b,t}$ is any scalar diagnostic for trajectory $b$ at time $t$,
$B$ is the number of trajectories supplied for the class, and
$\operatorname{Var}_t$ is sample variance across timestamps.
$\mathcal V$ therefore averages temporal variance across trajectories;
$\mathcal M$ averages every supplied entry; and $\mathcal M_t$ averages only
over time. The vector $\mathbf a_t$ is centered finite-difference
acceleration, $\mathbf v_t$ is velocity, and $h$ is the frame interval.
Within Table~\ref{tab:appendix-pis3d-formulas}, $\mathbf p$,
$\mathbf v$, $\boldsymbol\omega$, $\mathbf u$, and $\mathbf a$ are predicted
position, velocity, angular velocity, scale rate, and acceleration;
subscripts $x,y,z$ select Cartesian components; and
$s_xs_ys_z$ is the product of the three predicted scale components.
$\|\cdot\|$ denotes Euclidean norm. Superscripts $-$ and $+$ denote the
samples immediately before and after a detected vertical-velocity sign
change, and $\mathcal M_{\mathrm{sign\ flip}}$ averages only those events.
$e$ and $\mu$ are predicted restitution and tangential attenuation.
The constant $9.81$ is the gravity magnitude, $22^\circ$ is the fixed slope
angle, and $10^{-6}$ prevents division by zero. All absolute-value, ReLU,
sine, cosine, variance, and mean operations in the table have their standard
scalar meanings.

\subsection{Inference Procedure}

Beyond the forward path in the main paper, user-specified values can replace initial linear velocity, angular velocity, mass, restitution, and tangential attenuation. The time grid includes $t=0$, and the first returned state is the projected input state.

When cameras are unavailable for qualitative rendering, inference constructs equally spaced orbit views with a 55-degree pinhole field of view, radius 5, elevation 1.2, and target equal to the mean predicted position. The rendering implementation can use the CUDA \texttt{gsplat} rasterizer~\citep{ye2025gsplat} or the lightweight fallback. Neither renderer affects the state metrics.

\section{Dataset Construction}

This section defines four data resources in order. Gaussian-32 is the
appearance-conditioned Gaussian evaluation set; State-32 is the procedural
state-trajectory corpus; Same-32 is the fixed 32-sample cross-representation
subset; and Common-12 is the aligned 84-video manifest used by the video proxy
comparison.

\subsection{Gaussian-32}

Beyond the Gaussian-32 summary in the main paper, its executed construction used one sample per object-motion pair, 49 frames, $\Delta t=1/24$ seconds, hard-profile sampling, seed 2026, object scale $0.9$ with per-axis jitter $U(0.85,1.18)$, 1,536 object Gaussians, and 3,200 static background Gaussians. Thus each full scene contains 4,736 Gaussians. Supported objects are aligned to the plane with a clearance of $0.015$; airborne classes use class-specific heights.

The set stores four orbit views at $128\times128$. Cameras are evenly spaced in azimuth, have radius 5, elevation 1.35, and a 55-degree field of view, and look at the mean trajectory position plus the template's target-height offset. We use the lightweight renderer with 2D splat standard deviation $1.6$, a semantic 3D Gaussian background, and no 2D-flow export. Every sample contains color and RGB/depth/mask supervision.

In the fast no-flow lightweight renderer, camera-space $y$ is negated before
pixel projection. A square kernel of radius $\lceil2\sigma\rceil$ is evaluated
around the rounded center, where $\sigma=1.6$ pixels is the 2D Gaussian
standard deviation, the factor 2 truncates the footprint at two standard
deviations, and $\lceil\cdot\rceil$ rounds upward to an integer pixel radius.
Weights are the 2D Gaussian kernel times opacity and are discarded below
$10^{-4}$, the fixed contribution threshold. Pixel alpha, alpha-weighted
depth, and alpha-weighted color are accumulated with scatter-add; alpha is
clamped to $[0,1]$, and the remaining weight is filled with the specified
background color. This path does not use the 3D covariance when computing the
2D footprint, so it is a point-splat proxy rather than a full 3DGS rasterizer.

\begin{table*}[!ht]
\centering
\small
\begin{tabular}{lll}
\toprule
Scene identifier & Procedural object template & Scene description \\
\midrule
\texttt{living\_room\_table} & red toy car & wooden table in a bright living room \\
\texttt{studio\_tabletop} & blue ceramic mug & tabletop under soft studio lighting \\
\texttt{sunny\_park} & black-and-white soccer ball & short grass in a sunny park \\
\texttt{kitchen\_counter} & green glass bottle & counter with a warm wall \\
\texttt{orbit\_space} & cratered moon sphere & starry space scene \\
\texttt{playroom\_floor} & red cube & simple playroom floor \\
\texttt{desk\_blue\_sphere} & blue sphere & gray desktop \\
\texttt{construction\_floor} & yellow cone & simple concrete floor \\
\texttt{lab\_table\_cylinder} & purple cylinder & clean laboratory table \\
\texttt{display\_orange\_pyramid} & orange pyramid & simple display surface \\
\bottomrule
\end{tabular}
\caption{The ten procedural object-scene templates used by Gaussian-32. Geometry, color, and background Gaussians are generated procedurally; these entries are not downloaded third-party assets.}
\label{tab:appendix-gaussian32-templates}
\end{table*}

\paragraph{Gaussian-32 sample schema.}
Every sample is a PyTorch dictionary. The labeled state tensor has shape $(49,22)$. The canonical object contains \texttt{local\_means}, \texttt{colors}, \texttt{opacities}, and \texttt{covariances} with shapes $(1536,3)$, $(1536,3)$, $(1536)$, and $(1536,3,3)$. Static-background means, colors, opacities, and covariances have corresponding shapes with 3,200 primitives. The transformed object means and covariances have shapes $(49,1536,3)$ and $(49,1536,3,3)$; the composed scene uses $(49,4736,3)$ and $(49,4736,3,3)$. Object and full-scene 3D flows have shapes $(48,1536,3)$ and $(48,4736,3)$. Event metadata contains a collision indicator, contact normal, and impulse at each of the 49 timestamps.

Camera intrinsics, world-to-camera transforms, and camera-to-world transforms are stored as \texttt{K}, \texttt{w2c}, and \texttt{c2w} with shapes $(4,3,3)$, $(4,4,4)$, and $(4,4,4)$. Render supervision stores float32 RGB, depth, and mask tensors with shapes $(49,4,128,128,3)$, $(49,4,128,128)$, and $(49,4,128,128)$. Metadata records the sample and scene identifiers, motion type, prompts, object template, asset and background provenance, rendering mode, and initial physical controls.

\subsection{State-32}

We refer to the final state corpus as \emph{State-32}. It contains 32 motion labels: 3D uniform motion, gravity acceleration, free fall, projectile motion, airplane flight, helical flight, circular orbit, 3D rotation, size change, damped pendulum, slope sliding, rolling with friction, planar bouncing, wall collision, nonlinear force field, nonrigid deformation, hybrid collision impulse, figure-eight flight, decaying spiral orbit, damped bouncing, rolling followed by collision, sliding to a stop, throwing and landing, vertical launch, wind-drag projectile motion, spring oscillation, stop-and-go motion, two-stage motion, banked airplane turns, orbit with precession, tumbling fall, and scale pulse.

For the class-specific definitions below, $h=1/24$ seconds is the fixed time
step and $k\in\{0,\ldots,63\}$ is the zero-based frame index. We write
$U(a,b)$ for an independent sample from the continuous uniform distribution
on the closed interval with lower bound $a$ and upper bound $b$. Subscripts
$0$, $x$, $y$, and $z$ denote the initial frame and Cartesian components.
Thus $p_{0y}$ is initial vertical position, $s_{0y}$ is initial vertical
scale, and $v_{0y}$ is initial vertical velocity. The hard-profile base ranges
are summarized in Table~\ref{tab:appendix-sampling}. Airborne classes overwrite
the sampled height with $p_{0y}=\bar h_c+U(0,0.45)$, where $\bar h_c$ is the
class-specific nominal height: 3.4 (free fall), 1.8 (projectile and the default
airborne case), 1.9 (gravity acceleration and figure eight), 2.3 (airplane),
2.0 (helix), 1.6 (circular orbit and precession), 1.7 (decaying spiral), 1.2
(throw and land), 0.8 (vertical launch), 1.4 (wind projectile), 1.5 (spring),
2.1 (banked turn), or 3.0 (tumbling fall). Non-airborne classes use
$p_{0y}=s_{0y}+U(0,0.04)$. Projectile samples additionally enforce
$v_{0y}\geq2.2$. All numerical ranges in this subsection are in the synthetic
coordinate and time units of State-32.

\begin{table*}[!ht]
\centering
\small
\begin{tabular}{lll}
\toprule
Variable & Train and \texttt{val\_id} & \texttt{val\_ood} \\
\midrule
$p_x,p_z$ & $U(-0.8,0.8)$ & same \\
$v_x,v_z$ & $U(-3.2,3.2)$ & $U(-5.12,5.12)$ \\
$v_y$ before class rules & $U(0.4,4.8)$ & $U(0.4,7.68)$ \\
$\omega_x,\omega_y,\omega_z$ & $U(-4,4)$ & $U(-6.4,6.4)$ \\
$s_x,s_y,s_z$ & $U(0.18,0.36)$ & same \\
$m$ & $U(0.45,3.625)$ & same \\
$e$ & $U(0.35,0.96)$ & same \\
$\mu$ & $U(0.12,0.75)$ & same \\
\bottomrule
\end{tabular}
\caption{Base sampling ranges in the hard profile. The velocity-range-shift split expands the horizontal-velocity interval, the upper (but not lower) vertical-velocity bound, and the angular-velocity interval by 1.6; it does not hold out motion labels.}
\label{tab:appendix-sampling}
\end{table*}

\paragraph{Complete Procedural Generator Definition.}

The following specification records the executed State-32 generator rather
than an approximate physical interpretation. The sampled time is $t_k=kh$,
where $t_k$ is elapsed time at frame $k$. The fixed gravity vector is
$\mathbf g=(0,-9.81,0)$, and
$\bar s=(s_{0x}+s_{0y}+s_{0z})/3$ is the arithmetic mean of the three initial
scale components. For a vector sequence $\mathbf x_k$, the centered
finite-difference operator is
$D_h[\mathbf x]_k=(\mathbf x_{k+1}-\mathbf x_{k-1})/(2h)$ at interior frames,
with $(\mathbf x_1-\mathbf x_0)/h$ and
$(\mathbf x_{63}-\mathbf x_{62})/h$ at the first and final frames.
The forward operator is
$F_h[\mathbf x]_k=(\mathbf x_{k+1}-\mathbf x_k)/h$ for $k<63$, with the same
backward difference at $k=63$. Unless a row states otherwise,
$\mathbf s_k=\mathbf s_0$ and
$\boldsymbol\omega_k=\boldsymbol\omega_0$. The notation
$\operatorname{SI}(\mathbf a_k)$ means the semi-implicit recurrence
\begin{equation}
\mathbf v_{k+1}=\mathbf v_k+h\mathbf a_k,qquad
\mathbf p_{k+1}=\mathbf p_k+h\mathbf v_{k+1}.
\label{eq:data-si}
\end{equation}
In Eq.~\ref{eq:data-si}, $\mathbf p_k,\mathbf v_k,\mathbf a_k\in\mathbb R^3$
are position, velocity, and acceleration at frame $k$. The subscript $k+1$
denotes the next frame and $h$ is the fixed step defined above. The first
assignment updates velocity from acceleration; the second then uses that new
velocity $\mathbf v_{k+1}$ to update position, which is why the recurrence is
semi-implicit rather than fully explicit.

\paragraph{Notation for the generator tables.}
Bold symbols are three-dimensional vectors and ordinary italic symbols are
scalars unless stated otherwise. The subscripts $x,y,z$ select Cartesian
components, $0$ selects the sampled initial value, and $k$ selects a frame.
$\mathbf p,\mathbf v,\mathbf a,\boldsymbol\omega,\mathbf s$ denote position,
linear velocity, acceleration, angular velocity, and scale. The symbol
$\mathbf0$ is the zero vector; $\|\cdot\|$ is Euclidean norm;
$|\cdot|$ is scalar absolute value; $\odot$ is component-wise multiplication;
and $\max$ and $\min$ select the larger or smaller argument, component-wise
when a vector is present. The assignment arrow $\leftarrow$ means that the
generator overwrites the value on its left. The index $j$ in a sum is a
discrete frame index. The functions $\sin$, $\cos$, and $\exp$ are applied
component-wise to vectors, all trigonometric arguments are in radians, and
$\pi$ is the circle constant.

Symbols introduced within one motion row are local to that row. In particular,
$r$ or $\rho_k$ denotes an orbital radius, $w$ an angular frequency,
$\theta$ the slope angle, $a_\parallel$ acceleration along the slope,
$\mathbf d$ a unit travel direction, $d_k$ traveled distance, and $u_k$ scalar
speed. Depending on the named row, $a$ denotes a scalar acceleration or
oscillation amplitude, while $\mathbf a$ denotes a three-axis amplitude
vector. The symbols $y_f$, $e_b$, $\mathbf w$, $\boldsymbol\phi$, $P$, $b_k$,
$\mathbf c$, $\tau_k$, and $\nu$ denote the support height, current bounce
restitution, wind vector, phase-offset vector, stop-go period, binary motion
gate, second-stage orbit center, elapsed second-stage time, and precession
frequency. The symbol $\mathbb1[\cdot]$ is one when its bracketed condition is
true and zero otherwise. Every decimal coefficient in the tables is a fixed
procedural-generator constant, not a learned model parameter. These
definitions, Table~\ref{tab:appendix-sampling}, and the seeds below specify the
state targets without requiring a motion label at training time.

\begin{table*}[!ht]
\centering
\footnotesize
\setlength{\tabcolsep}{4pt}
\begin{tabular}{L{0.19\textwidth}L{0.75\textwidth}}
\toprule
Motion & Executed hard-profile definition \\
\midrule
3D uniform & $\mathbf p_k=\mathbf p_0+t_k\mathbf v_0$, $\mathbf v_k=\mathbf v_0$. \\
Gravity acceleration & $\mathbf a_k=[0.45+0.95\sin(2.3t_k),-4.2+1.2\cos(1.7t_k),0.30+0.80\sin(3.1t_k+0.4)]$; $\mathbf v_k=\mathbf v_0+h\sum_{j=0}^{k}\mathbf a_j$ and $\mathbf p_k=\mathbf p_0+h\sum_{j=0}^{k}\mathbf v_j$. \\
Projectile & $\mathbf p_k=\mathbf p_0+t_k\mathbf v_0+\tfrac12t_k^2\mathbf g$, $\mathbf v_k=\mathbf v_0+t_k\mathbf g$. \\
Circular orbit & $r=0.9+\bar s$, $w=1.2+|\omega_{0y}|$, $\rho_k=r[1+0.22\sin(2.1t_k)]$; $\mathbf p_k=[\rho_k\cos(wt_k),p_{0y}+0.35\sin(0.5wt_k),\rho_k\sin(wt_k)]$, $\mathbf v=D_h[\mathbf p]$, and $\boldsymbol\omega_k=[0.35\sin(2t_k),w,0.25\cos(1.5t_k)]$. \\
3D rotation & $\mathbf p_k=\mathbf p_0$, $\mathbf v_k=\mathbf0$, $\boldsymbol\omega_k=\boldsymbol\omega_0+[0,1.5,0]$. \\
Damped pendulum & $\mathbf p_k=\mathbf p_0+[1.15e^{-0.22t_k}\cos(4.2t_k),0.45e^{-0.22t_k}\sin(4.2t_k),0.35\sin(7.14t_k+0.3)]$; $\mathbf v=D_h[\mathbf p]$ and $\boldsymbol\omega_k=[0,0,4.2]e^{-0.22t_k}$. \\
Slope sliding & $\theta=22^\circ$, $a_\parallel=9.81\sin\theta-9.81\mu\cos\theta$, $d_k=v_{0x}t_k+\tfrac12a_\parallel t_k^2$; $\mathbf p_k=[p_{0x}+d_k\cos\theta,s_{0y}+d_k\sin\theta,p_{0z}]$ and $\mathbf v=D_h[\mathbf p]$. \\
Size changing & $\mathbf p_k=\mathbf p_0+0.75t_k\mathbf v_0$, $\mathbf v=D_h[\mathbf p]$; $\mathbf s_k=\mathbf s_0\odot[1+0.32\sin(3.1t_k),1+0.28\cos(2.3t_k+0.4),1+0.24\sin(4.5t_k+0.2)]$, with each factor lower-bounded by 0.35. \\
Nonrigid deformation & $\mathbf p_k=\mathbf p_0+0.65t_k\mathbf v_0$, $\mathbf v=D_h[\mathbf p]$; for $a_k=1+0.42\sin(5t_k)$, $\mathbf s_k=\mathbf s_0\odot[a_k,1/\max(a_k,0.2),1+0.18\cos(4t_k)]$. This is only aggregate anisotropic scaling. \\
\bottomrule
\end{tabular}
\caption{Closed-form generators inherited from the base hard-profile corpus.}
\label{tab:appendix-generators-a}
\end{table*}

\begin{table*}[!ht]
\centering
\footnotesize
\setlength{\tabcolsep}{4pt}
\begin{tabular}{L{0.19\textwidth}L{0.75\textwidth}}
\toprule
Motion & Executed hard-profile definition and event rule \\
\midrule
Planar bouncing & Initialize $p_y\leftarrow\max(p_y,s_{0y}+0.35)$ and $\mathbf v\leftarrow[v_{0x}+1.4,|v_{0y}|+2,v_{0z}-1.2]$. Use Eq.~\ref{eq:data-si} with $\mathbf a_k=\mathbf g+[0.8\sin(3.4t_k),0,0.6\cos(2.1t_k)]$. If $p_{k+1,y}<s_{0y}$ and $v_{k+1,y}<0$, set $p_{k+1,y}=s_{0y}$, $v_{k+1,y}=-e v_{k+1,y}$, and $(v_x,v_z)\leftarrow(1-\mu)(v_x,v_z)$. \\
Wall collision & Initialize $p_x=-1.8$ and $\mathbf v=[|v_{0x}|+2.2,0.2,0.8]$. At each step apply $v_y\leftarrow v_y-2h$ and $\mathbf p\leftarrow\mathbf p+h\mathbf v$. If $p_x+s_{0x}>1.1$ and $v_x>0$, set $p_x=1.1-s_{0x}$, $v_x=-ev_x$, and $v_z\leftarrow(1-\mu)v_z$. \\
Rolling with friction & $\mathbf d=[1,0,0.3]/\|[1,0,0.3]\|$, $u_0=|v_{0x}|+2$, $a=1.35(9.81)\mu$, $d_k=\max(0,u_0t_k-\tfrac12at_k^2)$, and $u_k=\max(0,u_0-at_k)$. Set $\mathbf p_k=\mathbf p_0+d_k\mathbf d$ with $p_y=s_{0y}$, $\mathbf v_k=u_k\mathbf d$, and $\boldsymbol\omega_k=[0,0,u_k/\bar s]$. \\
Hybrid collision & Initialize $p_y\leftarrow\max(p_y,s_{0y}+0.45)$ and $\mathbf v=[|v_{0x}|+1.8,|v_{0y}|+2.8,v_{0z}]$. Use Eq.~\ref{eq:data-si} with $\mathbf a_k=\mathbf g+[0.45\sin(5t_k),0,0.35\cos(3t_k)]$. Floor contact at $s_{0y}$ uses $v_y\leftarrow-ev_y$, $v_x\leftarrow(1-\mu)v_x$, $v_z\leftarrow(1-0.5\mu)v_z$. Wall contact at $p_x+s_{0x}>1.25$ uses $p_x=1.25-s_{0x}$, $v_x\leftarrow-ev_x$, $v_z\leftarrow v_z+0.6$. Here $\boldsymbol\omega_k=[0.7\sin(3t_k),|\omega_{0y}|+1,0.4\cos(2.1t_k)]$. \\
Nonlinear force field & Starting from $(\mathbf p_0,\mathbf v_0)$, use Eq.~\ref{eq:data-si} with $\mathbf a_k=[-1.15p_x,-0.2(p_y-1),-1.05p_z]+[0.85\sin(3.2t_k+p_z),-2.2,0.85\cos(2.7t_k+p_x)]-0.18\min(\|\mathbf v\|,8)\mathbf v$. Floor contact at $s_{0y}$ uses the planar-bounce rule. Scale factors are $[1+0.18\sin(3.7t),1+0.15\cos(2.9t),1+0.12\sin(4.5t+0.2)]$, lower-bounded by 0.5; $\boldsymbol\omega=\boldsymbol\omega_0+[0.5\sin(2t),0.5\cos(1.5t),0.4\sin(2.8t)]$. \\
\bottomrule
\end{tabular}
\caption{Base generators with stopping or collision events. The generated event tensors are metadata and are not model inputs.}
\label{tab:appendix-generators-b}
\end{table*}

For the remaining custom generators, the support coordinate is $y_f=\max(0.5s_{0y},0.04)$. The operator $\operatorname{StopFloor}$ finds the first sampled state with $p_y<y_f$, sets $p_y=y_f$ and $v_y=0$ from that sample onward, marks those samples as contacts, and stores one normal impulse $m|v_y|$ from the sample immediately before the first crossing. It does not bounce. This convention differs from the $s_{0y}$ support coordinate in the base collision generators above.

\begin{table*}[!ht]
\centering
\footnotesize
\setlength{\tabcolsep}{4pt}
\begin{tabular}{L{0.19\textwidth}L{0.75\textwidth}}
\toprule
Motion & Executed expanded-set definition \\
\midrule
Free fall & Replace $\mathbf v_0$ by $[0.25v_{0x},0,0.18v_{0z}]$, use the projectile equations, then apply $\operatorname{StopFloor}$. \\
Airplane & $u=1.6+0.35|v_{0x}|$; $\mathbf p_k=[p_{0x}+ut_k,p_{0y}+0.18\sin(2\pi t_k/t_{63}),p_{0z}+0.38\sin(1.6t_k)]$, $\mathbf v=F_h[\mathbf p]$, and $\boldsymbol\omega=[0,0.35,0]$. \\
Helical flight & $r=0.72+0.20\bar s$, $w=2.3+0.25|\omega_{0y}|$; $\mathbf p_k=[p_{0x}+r\cos(wt_k),p_{0y}+0.28t_k+0.16\sin(1.7wt_k),p_{0z}+r\sin(wt_k)]$, $\mathbf v=F_h[\mathbf p]$, $\boldsymbol\omega=[0.25,w,0.10]$. \\
Figure eight & $a=0.95+0.15\bar s$, $w=1.95+0.2|\omega_{0y}|$; $\mathbf p_k=[p_{0x}+a\sin(wt_k),p_{0y}+0.20\sin(0.7wt_k+0.4),p_{0z}+0.55a\sin(2wt_k)]$, $\mathbf v=D_h[\mathbf p]$, and $\boldsymbol\omega=[0.18\cos(wt_k),0.45,0.22\sin(1.4wt_k)]$. \\
Decaying spiral & $w=2.55+0.25|\omega_{0y}|$ and $r_k$ is linearly spaced from 1.25 to 0.35; $\mathbf p_k=[p_{0x}+r_k\cos(wt_k),p_{0y}+0.16\sin(1.2wt_k),p_{0z}+r_k\sin(wt_k)]$, $\mathbf v=D_h[\mathbf p]$, $\boldsymbol\omega=[0.10,w,0.18]$. \\
Throw and land & Set $\mathbf v=[1.15+0.3v_{0x},|v_{0y}|+3.4,0.55v_{0z}]$, use the projectile equations, and apply $\operatorname{StopFloor}$. After first hit $k_h$, let $\tau=t-t_{k_h}$ and horizontal $\mathbf u$ equal the pre-hit velocity with $u_y=0$; add $\tau\mathbf u e^{-2.2\mu\tau}$ to stored positions and set $\mathbf v=\mathbf u e^{-2.2\mu\tau}$. \\
Vertical launch & Set $\mathbf v=[0.15v_{0x},4.2+|v_{0y}|,0.12v_{0z}]$, use the projectile equations, apply $\operatorname{StopFloor}$, and set $\boldsymbol\omega=[0.35,0.10,0.20]$. \\
Tumbling fall & Set $\mathbf v=[0.45v_{0x},0.25|v_{0y}|,0.35v_{0z}]$, use the projectile equations, and apply $\operatorname{StopFloor}$. Set $\boldsymbol\omega_k=\boldsymbol\omega_0+[3.2+0.5\sin(2t_k),2.2+0.4\cos(1.7t_k),2.6+0.45\sin(2.4t_k)]$. \\
\bottomrule
\end{tabular}
\caption{Expanded flight and ballistic generators. The initial-height sampling stated above is applied before these equations.}
\label{tab:appendix-generators-c}
\end{table*}

\begin{table*}[!ht]
\centering
\footnotesize
\setlength{\tabcolsep}{4pt}
\begin{tabular}{L{0.19\textwidth}L{0.75\textwidth}}
\toprule
Motion & Executed expanded-set definition and event rule \\
\midrule
Damped bouncing & Initialize $p_y\leftarrow\max(p_y,y_f+1.55)$, $\mathbf v=[0.75v_{0x}+0.7,|v_{0y}|+2.2,0.45v_{0z}]$, and $e_b=\min(0.85,\max(0.25,e))$. Use Eq.~\ref{eq:data-si} with $\mathbf g$. At a floor crossing set $p_y=y_f$, $v_y=-e_bv_y$, $(v_x,v_z)\leftarrow(1-0.45\mu)(v_x,v_z)$, then $e_b\leftarrow0.76e_b$; set $v_y=0$ when $|v_y|<0.22$. Set $\boldsymbol\omega_k=\boldsymbol\omega_0+[0.35\sin(4t_k),0.20\cos(2.6t_k),0.50\sin(3.2t_k)]$. \\
Rolling then collision & Initialize $p_x=-1.65$, $\mathbf d=[1,0,0.18]/\|[1,0,0.18]\|$, and $\mathbf v=(1.65+0.25|v_{0x}|)\mathbf d$. Advance $\mathbf p\leftarrow\mathbf p+h\mathbf v$; at $p_x+s_{0x}>1.2$, set $p_x=1.2-s_{0x}$, $v_x=-ev_x$, and $v_z\leftarrow v_z+0.35$. Every step multiplies $\mathbf v$ by $\max(0,1-0.55\mu h)$; $\boldsymbol\omega=[0,0,\|\mathbf v\|/\max(\bar s,10^{-3})]$. \\
Sliding then stop & $\mathbf d=[1,0,-0.28]/\|[1,0,-0.28]\|$, $u_0=1.8+0.4|v_{0x}|$, $a=\max(0.35,1.4(9.81)\mu)$, $u_k=\max(0,u_0-at_k)$, and $d_k=\min[u_0^2/(2a),\max(0,u_0t_k-\tfrac12at_k^2)]$. Set $\mathbf p=\mathbf p_0+d_k\mathbf d$ with constant $p_y$, $\mathbf v=u_k\mathbf d$, and $\boldsymbol\omega=\mathbf0$. \\
Wind-drag projectile & Initialize $\mathbf v=[1.55+0.25v_{0x},|v_{0y}|+2.6,0.55+0.15v_{0z}]$, $\mathbf w=[0.65,0,0.35]$, and $d=0.20+0.08\mu$. Use Eq.~\ref{eq:data-si} with $\mathbf a_k=\mathbf g+[0.25+0.15\sin(2.4t_k)]\mathbf w-d\min(\|\mathbf v-\mathbf w\|,8)(\mathbf v-\mathbf w)/\max(m,10^{-4})$. At the floor set $p_y=y_f$, $v_y=-0.25ev_y$, and $(v_x,v_z)\leftarrow(1-\mu)(v_x,v_z)$. Set $\boldsymbol\omega_k=\boldsymbol\omega_0+[0.4\sin(2t_k),0.3\cos(1.7t_k),0.4\sin(2.5t_k+0.4)]$. \\
Spring oscillation & With $\mathbf a=[0.85,0.42,0.38]$, $\boldsymbol\phi=[0,0.8,1.6]$, and anchor $\mathbf p_0+[0,0.2,0]$, set $\mathbf p_k=\text{anchor}+e^{-0.35t_k}\mathbf a\odot\sin(4.2t_k+\boldsymbol\phi)$, $\mathbf v=D_h[\mathbf p]$, and $\boldsymbol\omega=[0.35\sin(4.2t),0.25\cos(4.2t),0.30\sin(2.94t)]$. \\
Stop and go & $P=t_{63}/3$, $b_k=\mathbb 1[\sin(2\pi t_k/P)>0]$, $u_k=[1.25+0.25\sin(3.5t_k)]b_k$, and $\mathbf d=[1,0,0.35]/\|[1,0,0.35]\|$. Set $\mathbf v_k=u_k\mathbf d$, $\mathbf p_k=\mathbf p_0+h\sum_{j=0}^{k}\mathbf v_j$ with constant $p_y$, and $\boldsymbol\omega=[0,0,u_k]$. \\
Two stage & For $k<32$, $\mathbf p_k=\mathbf p_0+t_k[1.05,0,0.25]$ and $\boldsymbol\omega=[0,0.25,0]$. Let $\mathbf c=\mathbf p_{31}+[0,0,0.55]$ and $\tau_k=t_k-t_{32}$; for $k\geq32$, $\mathbf p_k=[c_x+0.55\sin(2.4\tau_k),p_{31,y}+0.12\sin(3.6\tau_k),c_z+0.55\cos(2.4\tau_k)]$ and $\boldsymbol\omega=[0.35,2.4,0.18]$. Use $\mathbf v=D_h[\mathbf p]$. \\
Banked airplane turn & $r=1.10$, $w=1.65$; $\mathbf p_k=[p_{0x}+r\sin(wt_k),p_{0y}+0.16\sin(0.8wt_k),p_{0z}+r(1-\cos(wt_k))]$, $\mathbf v=D_h[\mathbf p]$, $\boldsymbol\omega=[0.55\sin(wt),w,-0.45\sin(wt)]$. \\
Orbit with precession & $r=0.95$, $w=2.15$, $\nu=0.55$; $\mathbf p_k=[p_{0x}+r\cos(wt_k)\cos(\nu t_k),p_{0y}+0.42\sin(wt_k),p_{0z}+r\sin(wt_k)+0.35\sin(\nu t_k)]$, $\mathbf v=D_h[\mathbf p]$, $\boldsymbol\omega=[0.20\sin(\nu t),w,0.25\cos(\nu t)]$. \\
Scale pulse & $\mathbf p_k=\mathbf p_0+[0.12\sin(1.5t_k),0,0.10\cos(1.2t_k)]$ and $\mathbf v=D_h[\mathbf p]$. Scale factors are $[1+0.38\sin(4t),1+0.28\sin(4t+1.2),1+0.33\cos(3.2t)]$, lower-bounded by 0.35; $\boldsymbol\omega=[0.10,0.45,0.15]$. \\
\bottomrule
\end{tabular}
\caption{Remaining expanded generators.}
\label{tab:appendix-generators-d}
\end{table*}

\paragraph{Event metadata.}
For each detected base floor contact, the generator marks the current loop index, stores normal $[0,1,0]$, and records normal impulse $-(1+e)mv_y^{-}$; wall events analogously use normal $[-1,0,0]$ and impulse $-(1+e)mv_x^{-}$. Hybrid contact uses the corresponding rule for whichever event fires, with a shared event flag if both occur in one step. Damped bouncing substitutes the current $e_b$ for $e$, rolling-then-collision uses the wall rule, and wind-drag contact uses the base floor impulse even though its executed post-impact velocity has the additional factor 0.25. The $\operatorname{StopFloor}$ metadata rule was defined above. These arrays document generator events, but only the post-event state trajectory is used as a training target.
Here $[0,1,0]$ and $[-1,0,0]$ are the inward unit normals of the floor and
wall; superscript ${}^{-}$ denotes the velocity immediately before impact;
$v_y^{-}$ and $v_x^{-}$ are its normal components; $m$ is mass; and $e$ is
normal restitution. The leading minus sign makes the stored scalar impulse
nonnegative for an incoming negative normal velocity, while $(1+e)$ accounts
for reversal and restitution.

\paragraph{State assembly.}
After constructing $(\mathbf p_k,\mathbf v_k,\boldsymbol\omega_k,\mathbf s_k)$, the generator sets $\mathbf u_k=D_h[\mathbf s]_k$ and
\begin{equation}
\mathbf q_k=\mathcal Q_{\mathrm{AA}}
\left(\frac{\boldsymbol\omega_k}{\|\boldsymbol\omega_k\|},
\|\boldsymbol\omega_k\|t_k\right),
\end{equation}
where $\mathbf q_k\in\mathbb R^4$ is the orientation quaternion at frame $k$.
The function $\mathcal Q_{\mathrm{AA}}(\mathbf n,\vartheta)$, implemented by
\path{AxisAngleToQuaternion}, converts a unit rotation axis $\mathbf n$ and
rotation angle $\vartheta$ to a scalar-first quaternion. Here
$\mathbf n=\boldsymbol\omega_k/\|\boldsymbol\omega_k\|$ is the normalized
angular-velocity direction and
$\vartheta=\|\boldsymbol\omega_k\|t_k$ is the accumulated angle implied by
holding its current magnitude constant from time zero to $t_k$.
$\|\boldsymbol\omega_k\|$ is Euclidean angular speed and $10^{-6}$ is the
threshold below which normalization is avoided. The identity quaternion
$[1,0,0,0]$ is used below that threshold. This constructs each orientation
from the current angular-velocity vector and absolute time; it does not
numerically integrate angular velocity. The sampled $(m,e,\mu)$ are copied to
all 64 states. Collision indicators, normals, and impulses are saved as
metadata but are not part of the 22 state channels and are not given to
NewtonGS.

\paragraph{Hidden-label mixture and identifiability.}
The dataset contains the motion index for analysis, but training and evaluation pass only $(Z_0,t_{1:T})$ to every predictor. Class $c$ has a uniform prior, but after observing the class-dependent initial state the actual task distribution is
\begin{equation}
p(Z_{1:T}\mid Z_0)=\sum_{c=1}^{32}p(c\mid Z_0)
p(Z_{1:T}\mid Z_0,c).
\end{equation}
Here $p(\cdot\mid\cdot)$ denotes a conditional probability distribution;
$Z_0$ is the observed initial 22-dimensional state; and
$Z_{1:T}=(Z_1,\ldots,Z_T)$ is the future state sequence. The discrete variable
$c\in\{1,\ldots,32\}$ is the hidden motion class. The factor $p(c\mid Z_0)$ is
the posterior probability of class $c$ after observing the initial state, and
$p(Z_{1:T}\mid Z_0,c)$ is that class's conditional trajectory distribution.
The summation marginalizes the unobserved class, producing the trajectory
distribution presented to a predictor that receives no class label. The
posterior weights need not equal the uniform prior $1/32$ because height rules,
initial velocities, and angular rates can correlate with $c$. Nevertheless,
the construction does not enforce disjoint supports of $Z_0$ across classes
and therefore does not prove that the generator is identifiable from $Z_0$.

Generation used 32 CPU shards. Shard $i\in\{0,\ldots,31\}$ used seed $7301+1000i$ for training, an offset of 10,000 for \texttt{val\_id}, and an offset of 20,000 for \texttt{val\_ood}. Each shard generated 1,024 training, 128 ID-test, and 128 OOD-test samples per motion; merging recovered the split sizes stated in the main paper. All tensors were checked for the expected shapes and finite values.

\paragraph{Serialization and split manifests.}
Each State-32 file is a PyTorch dictionary. Its float32 \texttt{states} entry has shape $(N,64,22)$, where $N$ is the corresponding split size reported in the main paper. The remaining entries record \texttt{state\_names}, \texttt{dt}, \texttt{profile}, \texttt{split}, per-sequence \texttt{motion\_types} and \texttt{motion\_indices}, the ordered list \texttt{motion\_type\_names}, and a generation note. The CSV manifest has one row per sequence and records the sample identifier, split, motion name and index, motion category, number of steps, time step, collision-event count, source shard, and source identifier. Motion names and indices are retained for stratified analysis but are not read by the predictor.

The executed corpus contains exactly three partitions named \texttt{train}, \texttt{val\_id}, and \texttt{val\_ood}. The \texttt{val} prefix is an implementation-level filename retained for compatibility: \texttt{val\_id} and \texttt{val\_ood} are exactly the ID and OOD test sets reported in the main paper, rather than validation subsets in addition to a missing test set. Both are generated independently of and are disjoint from the training set. Model selection uses the complete \texttt{val\_id} objective, and the reported ID and velocity-shift results are evaluations on \texttt{val\_id} and \texttt{val\_ood}, respectively; \texttt{val\_ood} is not used for checkpoint selection. The serialized state files occupy approximately 5.6\,GB, 706\,MB, and 706\,MB, respectively.

\subsection{Same-32}

Same-32 selects the
\texttt{construction\_floor}-\texttt{yellow\_cone} Gaussian-32 sample with
repeat index 0000 for each of the 32 ordered motion types. It therefore
contains exactly 32 samples. Appearance, scene, primitive counts, reference
view, and repeat index are fixed across the subset, while only the motion
family changes. The manifest records the exact sample identifier, motion type,
source path, and time interval $\Delta t$ for every row.

\subsection{Common-12}

The Common-12 manifest contains 84 videos: 12 categories for each of
NewtonGS, CogVideoX, PhyT2V, Sora, Veo3, Wan, and NewtonGen. NewtonGS uses
reference-camera view~0 from the seed-7301 Same-32 render. CogVideoX, Wan, and
NewtonGen are regenerated from the same 12 recorded prompts. PhyT2V, Sora, and
Veo3 use the corresponding videos distributed in the NewtonGen comparison
release; their prompt and random-seed metadata are absent from the release and
are therefore not reconstructed here.

CogVideoX uses \path{THUDM/CogVideoX-5b} at $720\times480$, 49 frames,
8\,fps, 30 inference steps, and guidance 6.0. Wan uses
\path{Wan2.1-T2V-1.3B-Diffusers} at $832\times480$, 49 frames, 16\,fps, 30
steps, and guidance 5.0. NewtonGen uses the CogVideoX-5B base with its released
\path{T2V5B_blendnorm_i18000_DATASET} LoRA, 30 steps, guidance 6.0,
degradation 0.5, and its selected noise-warp cartridge. For each generated
row, the actual generator seed is
\begin{equation}
\begin{split}
h_{\mathrm{row}}&=
\operatorname{SHA1}(\texttt{method|motion|prompt\_id}),\\
s_{\mathrm{row}}&=7301+\operatorname{int}(h_{\mathrm{row},1:8})
\bmod10^6.
\end{split}
\end{equation}
Here \texttt{method}, \texttt{motion}, and \texttt{prompt\_id} are the
manifest strings identifying the generator, semantic motion category, and
prompt. The vertical bar inside the typewriter string is a literal field
separator. $\operatorname{SHA1}$ maps the concatenated UTF-8 string to a
hexadecimal digest, and $h_{\mathrm{row}}$ denotes that digest for the current
manifest row. The slice $h_{\mathrm{row},1:8}$ contains its first eight
hexadecimal digits; $\operatorname{int}$ converts that substring to an
integer. The operation $\bmod 10^6$ keeps the integer remainder in
$\{0,\ldots,999999\}$, and 7301 is the fixed base-seed offset.
$s_{\mathrm{row}}$ is the resulting generator seed for that row. Because the
method name is hashed, numerically identical row seeds are not
forced across CogVideoX, Wan, and NewtonGen. The ``fixed seed'' statement means
that this deterministic rule and base seed are fixed. Released methods have no
known seed.

\begin{table*}[!ht]
\centering
\scriptsize
\setlength{\tabcolsep}{4pt}
\begin{tabular}{L{0.15\textwidth}L{0.20\textwidth}L{0.55\textwidth}}
\toprule
Common category & NewtonGS label & Recorded generated-method prompt \\
\midrule
Uniform motion & \path{3d_uniform_motion} & A small metal cube sliding steadily along a smooth laboratory bench, reflections visible on the surface, scattered tools in the background, captured from a fixed side camera. \\
Acceleration & \path{3d_acceleration_gravity} & A red sedan accelerating in a straight line on a clean highway, the road flat and clear, with only a pale sky and distant horizon in the background, captured from a fixed roadside camera. \\
Deceleration & \path{sliding_then_stop} & A red sedan brakes and decelerates in a straight line on a wet city street, with lights reflecting on the road and buildings in the background, captured by a fixed side-view camera. \\
Parabolic motion & \path{projectile_motion} & A single apple is thrown at an angle with an initial speed. The camera captures the motion from the side, showing the apple rising, reaching its peak, and then descending under gravity. The scene takes place in a bright open field under a clear blue sky, with soft sunlight casting gentle shadows on the ground. The background shows green grass and distant trees, adding depth and realism. \\
Circular motion & \path{circular_orbital_motion} & A comet with a glowing tail orbits a distant star along a stable circular path. A top-down perspective emphasizes the symmetrical orbit and the stationary central star. \\
3D motion & \path{airplane_flight} & A small metal cube slides from the distance along a laboratory bench towards the camera, reflections visible on the surface, scattered tools in the background, captured from a fixed oblique side camera. \\
Rotation & \path{3d_rotation} & A metal rod spinning on a concrete floor, faint scratches and dust visible, captured from a fixed top-down camera. \\
Parabolic motion with rotation & \path{tumbling_fall} & A thin cylindrical rod gently tossed, rotating along its long axis, fixed side camera, realistic reflections, ground shadows visible, subtle motion blur.A chalkboard eraser spinning while falling in a tilted arc, side camera captures, scattered chalk pieces in the classroom background, realistic shadows and lighting. \\
Slope sliding & \texttt{slope\_sliding} & A small metal cube sliding down a laboratory ramp, shiny reflections on its surface, scattered tools and wires in the background, captured from a fixed side camera parallel to the ramp. \\
Damped oscillation & \path{pendulum_damped_oscillation} & A realistic pendulum with a spherical bob swinging from a fixed pivot. The fixed camera captures the entire motion. \\
Size changing & \path{size_changing} & A red helium balloon gradually inflating in a sunny park, children playing in the background, trees casting soft shadows, captured from a stationary side camera. \\
Deformation & \path{non_rigid_deformation} & A piece of soft dough is evenly flattened on a workbench, captured by a fixed overhead camera. \\
\bottomrule
\end{tabular}
\caption{Fixed Common-12 mapping and prompt inventory. The prompt column is
used by CogVideoX, Wan, and NewtonGen. NewtonGS does not consume it, and prompt
metadata is unavailable for the three released-video methods. The stored
parabolic-with-rotation prompt concatenates two scene descriptions without a
separator; this is retained by the generator.}
\label{tab:appendix-common12-prompts}
\end{table*}

\FloatBarrier
\section{Cross-Representation and Video-Generation Evaluation Protocols}

Using the data resources defined in Section~B, this section specifies the
L4GM/Same-32 cross-representation comparison and the unmatched Common-12 video
proxy evaluation. It also records the constants used by the analytic state
baselines.

\subsection{L4GM/Same-32 Cross-Representation Protocol}

The executed L4GM comparison uses the official repository at commit
\texttt{b857d670bd}\allowbreak\texttt{a83a569e2b}\allowbreak
\texttt{ad774afe56}\allowbreak\texttt{483d7a8674} and the official
\path{recon.safetensors} and \path{interp.safetensors} checkpoints from
\path{jiawei011/L4GM}. Inference invokes \path{infer_4d.py} with mode
\texttt{big}; temporal
interpolation is disabled, although the interpolation checkpoint path remains
an argument required by the launcher.

For every Same-32 row, the input is the \emph{ground-truth} Gaussian-32
render-supervision video from reference view~0, not a NewtonGS prediction.
The 49 RGB frames are resized to $256\times256$, the stored object mask is
bilinearly resized, and the foreground is composited on white. The video is
encoded at 24\,fps. Four first-frame masked views are also exported to the
L4GM workspace using the same white composite. Samples are selected by sorting
the manifest by motion name and sample identifier and retaining one repeat per
motion; metric values do not enter selection.

Each normalized L4GM artifact contains 49 frames and 65,536 Gaussians, compared
with 1,536 dynamic object Gaussians in the target. The adapter preserves the
method's own Gaussian topology, canonicalizes field names, and applies the
opacity-weighted PCA lifter in this appendix independently at every frame.
Finite differences then provide velocity, angular velocity, and scale rate;
$(m,e,\mu)=(1,0.75,0.2)$ are inserted because L4GM does not predict these
channels.

Temporal correspondence is frame-index correspondence at $\Delta t=1/24$.
No Procrustes fit, center translation, axis permutation, unit rescaling, or
target-dependent spatial alignment is applied to the L4GM output before state
evaluation. Consequently the reported position, scale, and plane diagnostics
also reflect L4GM's reconstruction coordinate gauge. NewtonGS receives the
labeled target $Z_0$, whereas L4GM receives masked RGB video and reconstructs
its own topology. This is therefore an exploratory interface check rather than
a controlled same-input dynamics comparison. The single deterministic set of
32 L4GM predictions is reused beside all three NewtonGS seeds; the displayed
zero L4GM standard deviation does not represent three independent L4GM runs.

\subsection{Common-12 Proxy Evaluation Protocol}

The evaluator converts every video to RGB in $[0,1]$, uniformly subsamples to
at most 32 frames, and downsamples only when the long edge exceeds 256 pixels.
It does not align objects, cameras, duration, frame rate, crop, scene, or
appearance. Thus the mapping in Table~\ref{tab:appendix-common12-prompts} is
semantic category alignment rather than matched input.

Let $\delta_b$ be the mean absolute consecutive-frame difference in the outer 8\% image border and $\delta_a$ the mean absolute second temporal difference over the full image. The reported background and smoothness proxies are
\begin{equation}
{\rm BC}_p=\exp(-12\delta_b),\qquad
{\rm MS}_p=\exp(-10\delta_a).
\end{equation}
Here ${\rm BC}_p$ is the background-consistency proxy and ${\rm MS}_p$ is the
motion-smoothness proxy; subscript $p$ marks both quantities as proxy metrics
rather than official benchmark scores. $\delta_b\geq0$ is the mean absolute
RGB difference between consecutive frames restricted to the outer 8\% image
border, while $\delta_a\geq0$ is the mean absolute second temporal difference
over all RGB pixels. The function $\exp$ is the natural exponential.
The fixed sensitivities 12 and 10 determine how quickly each score decays:
zero difference produces score one and larger difference approaches zero.

The normalized motion term is
$D=\operatorname{clip}(\delta_t/0.05,0,1)$, where $\delta_t$ is mean
full-image consecutive-frame L1 difference, $0.05$ is the fixed normalization
scale, and $\operatorname{clip}(x,0,1)=\min(1,\max(0,x))$ bounds the result.
Subject consistency $S$ is the mean similarity between the first frame and
later frames, computed as one minus the Bhattacharyya distance between
24-by-16 HSV histograms over the centered 55\% crop. The composite score is
\begin{equation}
{\rm PIS}_p=0.35{\rm MS}_p+0.25{\rm BC}_p+0.20D+0.20S.
\end{equation}
In this expression, ${\rm PIS}_p$ is the composite video proxy;
${\rm MS}_p$, ${\rm BC}_p$, $D$, and $S$ are the smoothness, background,
motion-magnitude, and subject-consistency terms defined above. Their fixed
weights $0.35$, $0.25$, $0.20$, and $0.20$ sum to one, so the composite is a
convex weighted average when all four inputs lie in $[0,1]$.
The table reports the arithmetic mean and sample standard deviation over the
12 categories. These image-statistic proxies are not official metrics of any
listed generator. NewtonGS is conditioned on a Gaussian scene and physical
state, generated methods are conditioned on the prompt inventory above, and
released methods use undocumented release-time conditions. Accordingly the
Common-12 table is an unmatched qualitative proxy inventory, not evidence for
a controlled cross-method quality ranking.

\subsection{Analytic-Baseline Constants}

All baselines receive the same projected labeled $Z_0$ and preserve
$(m,e,\mu)$. Hold-$Z_0$ repeats all 22 initial channels. Const-Vel-SE(3) uses
\begin{equation}
\begin{aligned}
\mathbf p_t&=\mathbf p_0+t\mathbf v_0,&
\mathbf v_t&=\mathbf v_0,\\
\mathbf s_t&=\max(10^{-4},\mathbf s_0+t\mathbf u_0),&
\mathbf u_t&=\mathbf u_0.
\end{aligned}
\end{equation}
Here $t\geq0$ is elapsed time; subscripts $0$ and $t$ denote initial and
current quantities. The vectors $\mathbf p,\mathbf v,\mathbf s,\mathbf u
\in\mathbb R^3$ are position, velocity, scale, and scale rate.
The first line integrates constant linear velocity. The second line integrates
constant scale rate and then applies $\max(10^{-4},\cdot)$ component-wise, with
$10^{-4}$ serving as the minimum valid scale. The baseline orientation is
$\mathbf q_t=\mathbf q_0\otimes
\operatorname{AxisAngle}(\boldsymbol\omega_0/\|\boldsymbol\omega_0\|,
t\|\boldsymbol\omega_0\|)$ with constant $\boldsymbol\omega_0$.
Here $\mathbf q_0$ and $\mathbf q_t$ are initial and current unit
quaternions, $\boldsymbol\omega_0$ is constant initial angular velocity,
$\|\boldsymbol\omega_0\|$ is angular speed,
$\boldsymbol\omega_0/\|\boldsymbol\omega_0\|$ is the rotation axis,
$t\|\boldsymbol\omega_0\|$ is the angle, $\operatorname{AxisAngle}$ converts
that axis-angle pair to a quaternion, and $\otimes$ is Hamilton
multiplication.

Damped-Vel-SE(3) sets $d(t)=\exp(-0.35t)$ and
$a(t)=(1-d(t))/0.35$, replacing $t$ above by $a(t)$ and multiplying
$\mathbf v_0,\boldsymbol\omega_0,\mathbf u_0$ by $d(t)$. Its quaternion angle
is $a(t)\|\boldsymbol\omega_0\|$.
The function $d(t)$ is the remaining-velocity fraction under damping rate
$0.35$, and $a(t)$ is its time integral, i.e., the effective displacement-time
coefficient. The natural exponential is denoted by $\exp$.

Gravity-Bounce-SE(3) and Physics-Prior-SE(3) use semi-implicit Euler once per
dataset frame, not RK4 substeps. At each frame,
\begin{equation}
\begin{aligned}
\mathbf v&\leftarrow\mathbf v+h\left(\mathbf g-c_v\mathbf v/m\right),\\
\mathbf p&\leftarrow\mathbf p+h\mathbf v,\\
\boldsymbol\omega&\leftarrow
\max(0,1-c_\omega h)\boldsymbol\omega,\\
\mathbf q&\leftarrow\mathbf q\otimes
\operatorname{AxisAngle}(\boldsymbol\omega,h\|\boldsymbol\omega\|).
\end{aligned}
\end{equation}
In this semi-implicit update, $\leftarrow$ means in-place assignment,
$h=1/24$ seconds is one dataset-frame step, and
$\mathbf g=(0,-9.81,0)$ is gravity. The vectors
$\mathbf v,\mathbf p,\boldsymbol\omega,\mathbf q$ are current linear velocity,
position, angular velocity, and orientation. The scalar $m$ is mass;
$c_v$ and $c_\omega$ are linear and angular damping constants.
$c_v\mathbf v/m$ is mass-normalized linear damping, while
$\max(0,1-c_\omega h)$ is a nonnegative per-step angular-velocity retention
factor. $\operatorname{AxisAngle}(\boldsymbol\omega,
h\|\boldsymbol\omega\|)$ uses the direction of
$\boldsymbol\omega$ as its axis and $h\|\boldsymbol\omega\|$ as its
incremental rotation angle; $\otimes$ composes this increment with the current
quaternion.
If scale restoration is active, it then applies
$\mathbf u\leftarrow\mathbf u+h[-0.25(\mathbf s-\mathbf1)-0.08\mathbf u]$
and $\mathbf s\leftarrow\max(10^{-4},\mathbf s+h\mathbf u)$; otherwise
$\mathbf s\leftarrow\max(10^{-4},\mathbf s+h\mathbf u)$. The same
horizontal-floor hit test and analytic restitution/tangential response as
NewtonGS follow the update, without a learned impulse. Gravity-Bounce uses
$c_v=c_\omega=0$ and no scale restoration. Physics-Prior uses
$c_v=c_\omega=0.05$ and scale restoration. None of the analytic baselines
receives the hidden motion label or future target states.
In these scale updates, $\mathbf s$ is scale, $\mathbf u$ is scale rate,
$\mathbf1=(1,1,1)$ is unit-scale equilibrium, $0.25$ is the restoration
coefficient, $0.08$ is scale-rate damping, and the component-wise lower bound
$10^{-4}$ prevents nonpositive scale.

\section{Ablation Analysis}

This section presents the residual diagnostics and clarifies the execution
status of additional ablation variants.

\subsection{Residual Analysis}

\paragraph{Initialization Comparison.}

Unless stated otherwise, the checkpoint-specific diagnostics from this subsection through the PIS-3D analysis use the seed-7301 reference checkpoint.

We compare the corrected zero-last model with the legacy zero-all checkpoint. Both use State-32, seed 7301, the same objective, hidden width, learning-rate schedule, and 100-epoch budget. The distributed batch schedules differ slightly: the legacy run used eight processes before resuming with six, whereas the corrected run uses seven processes throughout. There is one run per setting, so this comparison diagnoses the initialization defect but does not estimate seed variation.

\begin{table*}[!ht]
\centering
\small
\begin{tabular}{lrrrrr}
\toprule
Split & Traj & FDE & Vel & Quat & Scale \\
\midrule
ID  & 3.03\% & 5.50\% & 1.45\% & 5.18\% & 17.54\% \\
OOD & 2.33\% & 4.71\% & 1.40\% & 3.08\% & 18.40\% \\
\bottomrule
\end{tabular}
\caption{Relative error reduction from the legacy zero-all checkpoint to the corrected zero-last checkpoint. These are two single-run observations, not confidence intervals.}
\label{tab:appendix-init-change}
\end{table*}

\begin{figure*}[!ht]
\centering
\includegraphics[width=0.88\textwidth]{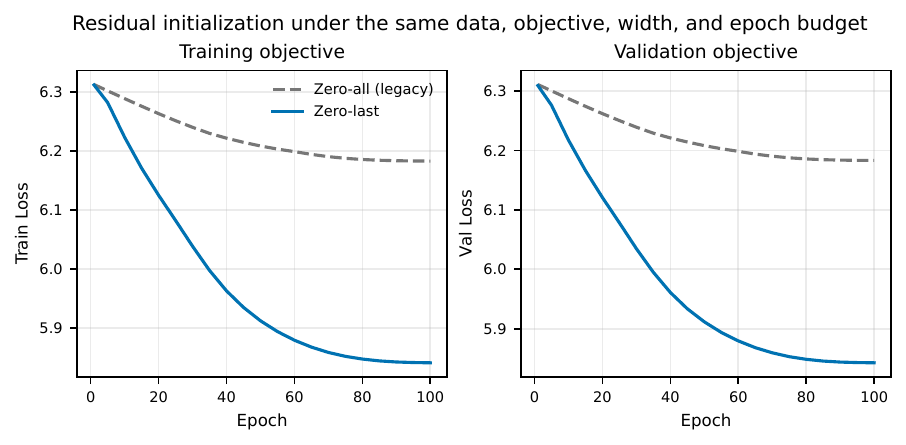}
\vspace{0.5em}
\includegraphics[width=0.88\textwidth]{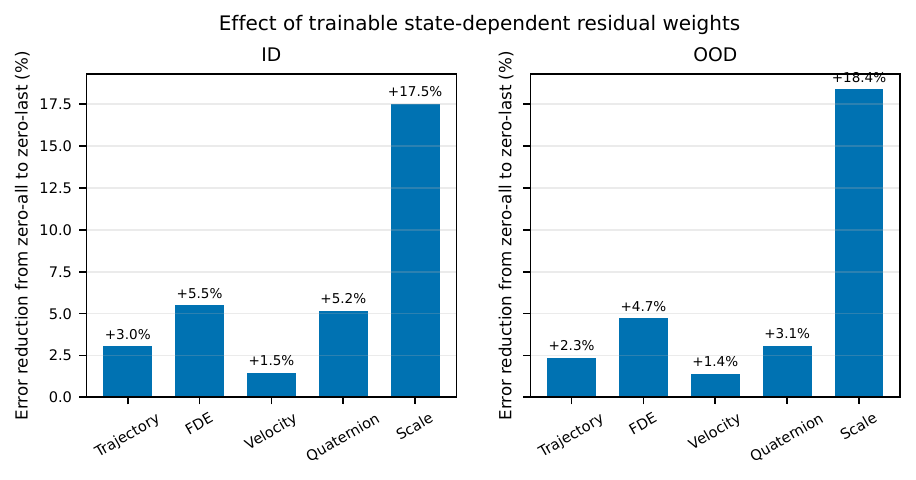}
\caption{Zero-all versus zero-last initialization. Top: recorded training and full-ID objectives. Bottom: relative changes in the five reported prediction errors. Both variants use one run; their distributed batch schedules are not identical.}
\label{fig:appendix-init}
\end{figure*}

\FloatBarrier
The final recorded ID objective decreases from 6.183299 to 5.842536.
Table~\ref{tab:appendix-init-change} reports changes in the complete-split
metrics; every listed error is lower with zero-last initialization.
Figure~\ref{fig:appendix-init} shows the objective curves and relative error
changes. Each table entry is
$100(E_{\mathrm{zero\mbox{-}all}}-E_{\mathrm{zero\mbox{-}last}})
/E_{\mathrm{zero\mbox{-}all}}$, where
$E_{\mathrm{zero\mbox{-}all}}$ is the error of the legacy initialization and
$E_{\mathrm{zero\mbox{-}last}}$ is the error of the corrected initialization.
The subtraction gives the error decrease, division makes it relative to the
legacy value, and multiplication by 100 expresses it as a percentage.

\paragraph{Post-Hoc Branch Removal.}

To measure whether the trained branches affect the selected checkpoint, we set
the continuous-residual parameters, contact-residual parameters, or both to
zero at evaluation time. We do not retrain the remaining model, so
Table~\ref{tab:appendix-branches} measures reliance on the learned branches
rather than the performance of independently optimized ablations.

\begin{table*}[!ht]
\centering
\resizebox{\textwidth}{!}{%
\begin{tabular}{llrrrrr}
\toprule
Split & Variant & Traj & FDE & Vel & Quat & Scale \\
\midrule
ID & Full & 3.167391 & 3.859234 & 3.648077 & 0.195692 & 0.163042 \\
ID & No continuous residual & 3.229521 & 4.020150 & 3.678065 & 0.209841 & 0.229285 \\
ID & No contact residual & 3.210038 & 3.945315 & 3.677260 & 0.195715 & 0.160706 \\
ID & No residual branches & 3.271227 & 4.100753 & 3.705204 & 0.208083 & 0.229285 \\
\midrule
OOD & Full & 3.738382 & 4.577842 & 4.006419 & 0.202348 & 0.161367 \\
OOD & No continuous residual & 3.790889 & 4.729918 & 4.042617 & 0.215592 & 0.229312 \\
OOD & No contact residual & 3.779623 & 4.668812 & 4.033973 & 0.199591 & 0.159074 \\
OOD & No residual branches & 3.831341 & 4.818889 & 4.068905 & 0.210234 & 0.229312 \\
\bottomrule
\end{tabular}%
}
\caption{Checkpoint branch removal on the complete ID and OOD splits. Lower is better. The variants are evaluated without retraining.}
\label{tab:appendix-branches}
\end{table*}

\FloatBarrier
Removing both branches increases trajectory RMSE by 3.28\% on ID and 2.49\%
on OOD; FDE increases by 6.26\% and 5.27\%, respectively. Removing either
branch alone also increases all three translation metrics, although removing
the contact branch slightly lowers scale RMSE and the OOD quaternion
discrepancy.

\subsection{Coverage and Execution Status}

\begin{table*}[!ht]
\centering
\small
\begin{tabular}{L{0.29\textwidth}L{0.19\textwidth}L{0.40\textwidth}}
\toprule
Requested variant & Evidence available & Interpretation \\
\midrule
Zero-all vs.\ zero-last initialization & Two single runs & Table~\ref{tab:appendix-init-change}; schedules are not identical and no seed variance is available. \\
Remove continuous residual & Post-hoc, full splits & Table~\ref{tab:appendix-branches}; selected checkpoint is not retrained. \\
Remove contact residual & Post-hoc, full splits & Table~\ref{tab:appendix-branches}; selected checkpoint is not retrained. \\
Remove both residuals & Post-hoc, full splits & Equivalent to evaluating the learned scalar analytic branch after zeroing both networks, not a separately optimized prior. \\
Disable contact event & Evaluator/model flag only & No matched 100-epoch retrained result is stored. \\
Remove analytic physics prior & No matched artifact & Requires a new model definition and retraining. \\
Euler or changed RK4 step size & No matched artifact & \texttt{max\_dt} exists in the model constructor, but the training launcher does not expose a reported sweep. \\
Remove individual state, penetration, or smoothness losses & Weight flags/code support only & No matched 100-epoch retrained checkpoints are stored. \\
\bottomrule
\end{tabular}
\caption{Status of the requested ablations. Numerical claims are limited to
stored full-split evaluations and the single-run initialization comparison.}
\label{tab:appendix-ablation-status}
\end{table*}

\FloatBarrier
Table~\ref{tab:appendix-ablation-status} separates reported evidence from
variants that the code can express but for which no matched run artifact
exists. This prevents an inference-time switch from being presented as a
retrained ablation. The repository also contains compact development/smoke
ablations with different widths, epoch budgets, and checkpoints; they are not
comparable to the three reported 100-epoch runs and are excluded.

\section{Additional Experimental Results}

This section reports motion-wise results and physical-parameter and metric
diagnostics.

\subsection{Motion-Wise Results}

The signed gain in Table~\ref{tab:appendix-per-motion} is
$100(E_{\mathrm{prior}}-E_{\mathrm{ours}})/E_{\mathrm{prior}}$; positive values
favor NewtonGS. These are seed-7301 class aggregates over all 4,096 samples per
class.
Here $E_{\mathrm{prior}}$ is the class-level trajectory RMSE of
Physics-Prior-SE(3), $E_{\mathrm{ours}}$ is the corresponding NewtonGS RMSE,
their difference is the absolute error reduction, division by
$E_{\mathrm{prior}}$ normalizes it relative to the baseline, and multiplication
by 100 expresses the result as a percentage.

\begin{table*}[!ht]
\centering
\scriptsize
\setlength{\tabcolsep}{3.5pt}
\begin{tabular}{lrrr|rrr}
\toprule
& \multicolumn{3}{c}{ID} & \multicolumn{3}{c}{OOD} \\
Motion & NewtonGS & Prior & Gain \% & NewtonGS & Prior & Gain \% \\
\midrule
3d\_acceleration\_gravity & 3.564 & 3.657 & +2.5 & 4.202 & 4.264 & +1.5 \\
3d\_rotation & 0.437 & 0.233 & -87.1 & 0.452 & 0.233 & -93.5 \\
3d\_uniform\_motion & 4.919 & 4.903 & -0.3 & 7.298 & 7.325 & +0.4 \\
airplane\_flight & 2.111 & 2.104 & -0.3 & 2.307 & 2.258 & -2.2 \\
banked\_airplane\_turn & 2.558 & 2.724 & +6.1 & 2.558 & 2.727 & +6.2 \\
bouncing\_on\_plane & 0.406 & 0.464 & +12.5 & 0.548 & 0.539 & -1.7 \\
circular\_orbital\_motion & 3.975 & 4.319 & +8.0 & 5.352 & 5.689 & +5.9 \\
damped\_bouncing & 0.453 & 0.486 & +6.7 & 0.579 & 0.569 & -1.7 \\
figure\_eight\_flight & 3.342 & 3.727 & +10.3 & 3.601 & 3.994 & +9.8 \\
free\_fall & 0.979 & 1.028 & +4.8 & 1.054 & 1.100 & +4.1 \\
helical\_flight & 2.689 & 2.946 & +8.7 & 2.834 & 3.097 & +8.5 \\
hybrid\_collision\_impulse & 5.494 & 5.715 & +3.9 & 8.294 & 8.549 & +3.0 \\
non\_rigid\_deformation & 3.384 & 3.437 & +1.5 & 5.016 & 5.089 & +1.4 \\
nonlinear\_force\_field & 2.272 & 2.281 & +0.4 & 3.825 & 3.879 & +1.4 \\
object\_wall\_collision & 3.515 & 3.705 & +5.1 & 4.164 & 4.279 & +2.7 \\
orbit\_with\_precession & 2.256 & 2.581 & +12.6 & 2.254 & 2.579 & +12.6 \\
pendulum\_damped\_oscillation & 2.563 & 2.812 & +8.8 & 2.563 & 2.812 & +8.9 \\
projectile\_motion & 10.282 & 10.392 & +1.1 & 9.023 & 9.114 & +1.0 \\
rolling\_then\_collision & 1.280 & 1.624 & +21.2 & 1.262 & 1.594 & +20.8 \\
rolling\_with\_friction & 0.942 & 0.984 & +4.3 & 1.759 & 1.803 & +2.5 \\
scale\_pulse & 0.450 & 0.390 & -15.4 & 0.452 & 0.390 & -15.9 \\
size\_changing & 3.853 & 3.839 & -0.4 & 5.807 & 5.821 & +0.2 \\
sliding\_then\_stop & 0.647 & 0.424 & -52.6 & 0.693 & 0.542 & -27.9 \\
slope\_sliding & 3.619 & 3.639 & +0.5 & 4.583 & 4.656 & +1.6 \\
spiral\_orbit\_decay & 3.506 & 3.824 & +8.3 & 3.817 & 4.139 & +7.8 \\
spring\_oscillation & 3.394 & 3.664 & +7.4 & 3.394 & 3.665 & +7.4 \\
stop\_and\_go\_motion & 0.839 & 1.153 & +27.2 & 0.839 & 1.153 & +27.2 \\
throw\_and\_land & 1.174 & 1.212 & +3.2 & 1.273 & 1.327 & +4.1 \\
tumbling\_fall & 0.998 & 1.040 & +4.0 & 1.190 & 1.224 & +2.8 \\
two\_stage\_motion & 1.136 & 1.379 & +17.7 & 1.138 & 1.381 & +17.6 \\
vertical\_launch & 0.805 & 0.855 & +5.8 & 0.810 & 0.893 & +9.2 \\
wind\_drag\_projectile & 0.976 & 0.959 & -1.7 & 1.409 & 1.393 & -1.1 \\
\bottomrule
\end{tabular}
\caption{Complete motion-wise trajectory RMSE for NewtonGS and
Physics-Prior-SE(3). ``Prior'' is one fixed method rather than a post-hoc
per-class best-baseline envelope.}
\label{tab:appendix-per-motion}
\end{table*}

\begin{figure*}[!ht]
\centering
\includegraphics[height=0.82\textheight,keepaspectratio]{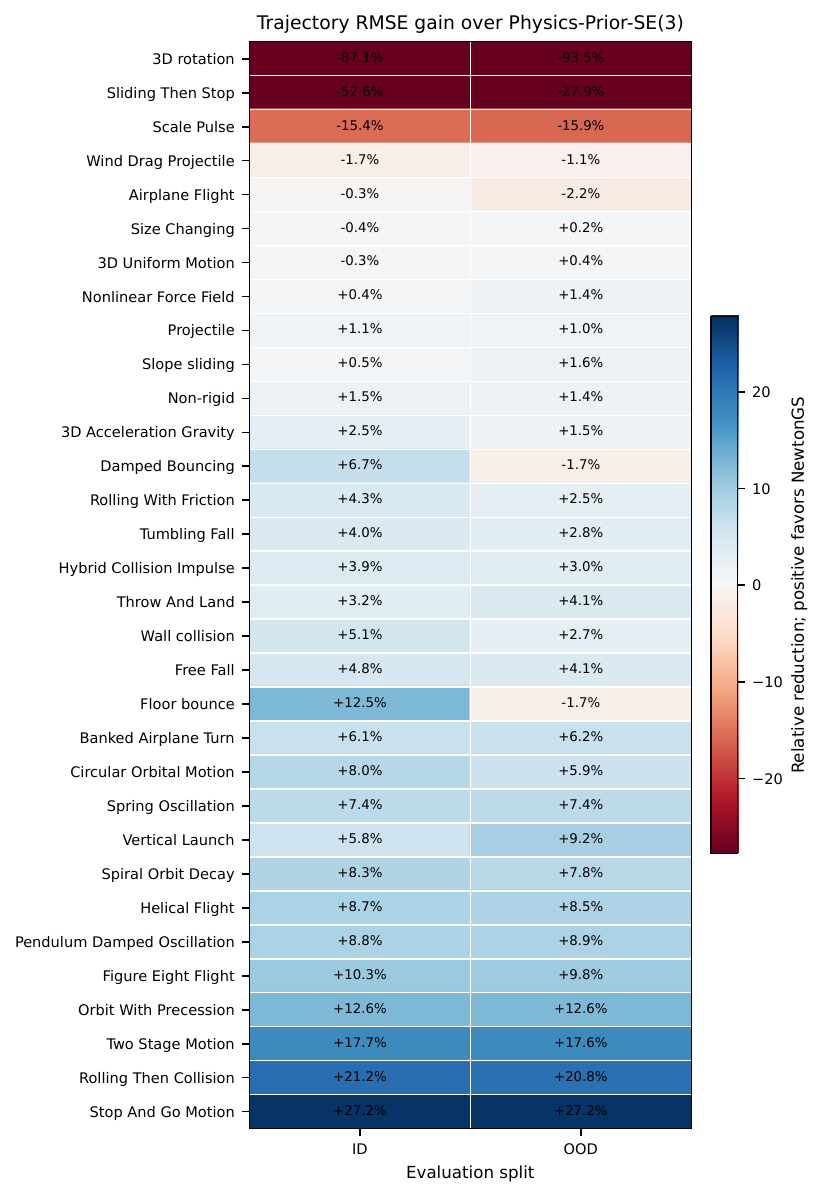}
\caption{Per-motion trajectory-RMSE reduction relative to Physics-Prior-SE(3). Positive values favor NewtonGS. Colors are clipped for readability, while the printed values are not clipped.}
\label{fig:appendix-motion-heatmap}
\end{figure*}

\FloatBarrier
Relative to Physics-Prior-SE(3), NewtonGS has lower trajectory RMSE on 25 of
32 motion classes in both splits. The median class-wise reduction is 4.52\% on
ID and 2.75\% on OOD. The largest consistent reductions occur for stop-and-go
motion, rolling followed by collision, and two-stage motion. The principal
failures are also consistent: pure 3D rotation, sliding to a stop, and scale
pulse have higher trajectory RMSE than the physics prior. In particular, the
large relative percentage for 3D rotation arises because its target
translation is small; the learned residual introduces position drift.
Figure~\ref{fig:appendix-motion-heatmap} reports every class instead of hiding
these failures in an aggregate mean.

\begin{table*}[!ht]
\centering
\small
\begin{tabular}{lrr|lrr}
\toprule
Motion & Global ID & Traj.\ RMSE & Motion & Global ID & Traj.\ RMSE \\
\midrule
Stop and go & 48,450 & 0.216858 & Pendulum & 13,532 & 2.113756 \\
Rolling then collision & 80,410 & 0.421288 & Helical flight & 58,102 & 2.043598 \\
Two stage & 73,180 & 0.453353 & Spiral decay & 76,110 & 2.325643 \\
Orbit with precession & 69,297 & 1.782550 & Circular orbit & 94,981 & 1.846519 \\
Planar bounce & 38,416 & 0.082722 & Spring oscillation & 113,837 & 2.490296 \\
Figure eight & 67,718 & 2.218789 & Damped bouncing & 84,369 & 0.126762 \\
\bottomrule
\end{tabular}
\caption{Exact sample identifiers for the trajectory figures. Every plot uses
all 64 timestamps, $k=0,\ldots,63$, and directly plots stored state
coordinates; there is no frame crop, color adjustment, or visual replacement.}
\label{tab:appendix-best-sample-ids}
\end{table*}

\begin{figure*}[!ht]
\centering
\includegraphics[width=\textwidth]{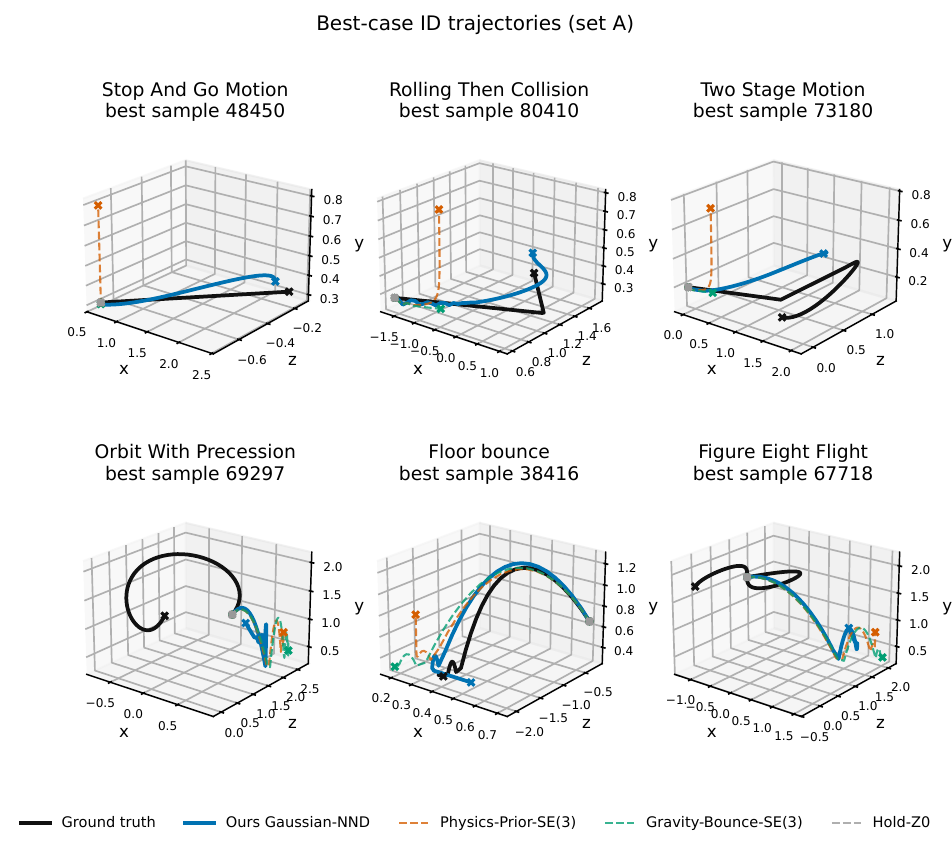}
\caption{ID trajectory visualizations, set A: stop-and-go motion, rolling then
collision, two-stage motion, orbit with precession, floor bounce, and
figure-eight flight. Dots and crosses mark initial and final states.}
\label{fig:appendix-best-trajectories-a}
\end{figure*}

\begin{figure*}[!ht]
\centering
\includegraphics[width=\textwidth]{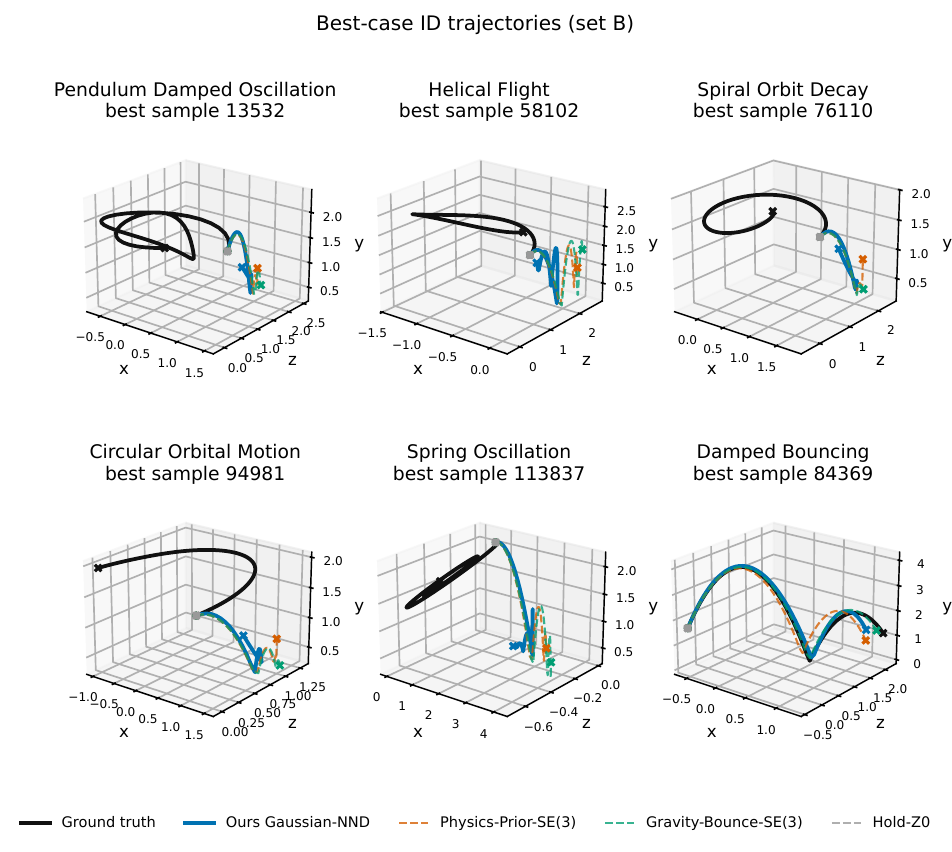}
\caption{ID trajectory visualizations, set B: damped pendulum, helical flight,
spiral-orbit decay, circular-orbital motion, spring oscillation, and damped
bouncing.}
\label{fig:appendix-best-trajectories-b}
\end{figure*}

\FloatBarrier
Figures~\ref{fig:appendix-best-trajectories-a}
and~\ref{fig:appendix-best-trajectories-b} visualize ID trajectories for 12
motion classes. Table~\ref{tab:appendix-best-sample-ids} records the global
sample identifier and trajectory RMSE for every panel. Each panel plots all 64
timestamps directly from the stored state coordinates.

\subsection{Physical-Parameter Response and Metric Diagnostics}

\subsubsection{Response to Explicit Physical Parameters}

\begin{figure*}[!ht]
\centering
\includegraphics[width=\textwidth]{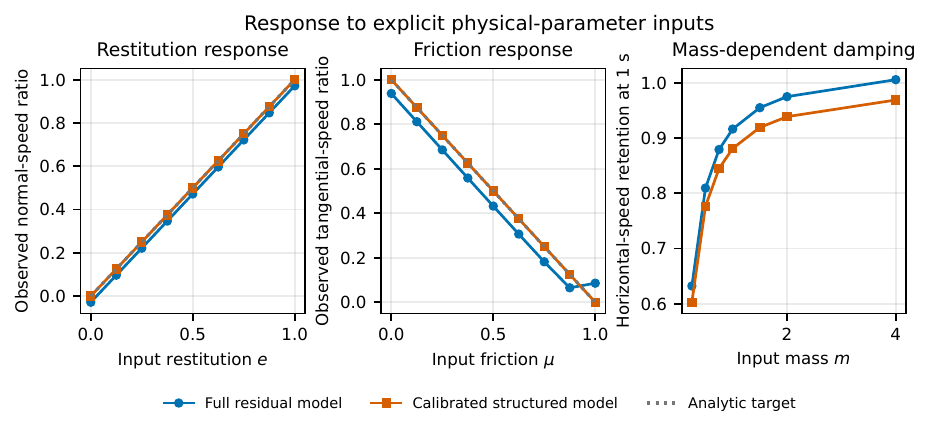}
\caption{Response to explicit restitution, tangential-attenuation, and mass inputs. The calibrated structured model zeros both learned residual branches while retaining the trained scalar coefficients. Dotted lines show the analytic restitution and attenuation targets.}
\label{fig:appendix-parameter-sweeps}
\end{figure*}

\FloatBarrier
Figure~\ref{fig:appendix-parameter-sweeps} varies one supplied parameter at a
time in fixed synthetic probes. The
observed normal-speed ratio is monotonic in restitution (Spearman $1.00$),
tangential-speed retention is nearly monotonic in the attenuation parameter
($-0.983$), and one-second horizontal-speed retention is monotonic in mass
($1.00$). These probes show that the inputs influence the implemented
response, but they do not test parameter estimation or identify ground-truth
materials. The learned residual also perturbs the analytic endpoints: at
$e=0$ the measured normal ratio is $-0.0284$, at $\mu=1$ tangential retention
is $0.0855$, and at $m=4$ horizontal-speed retention is $1.0056$. We therefore
describe the checkpoint as parameter responsive rather than perfectly
calibrated or physically guaranteed.

\subsubsection{PIS-3D Diagnostic}

\begin{figure*}[!ht]
\centering
\includegraphics[width=\textwidth]{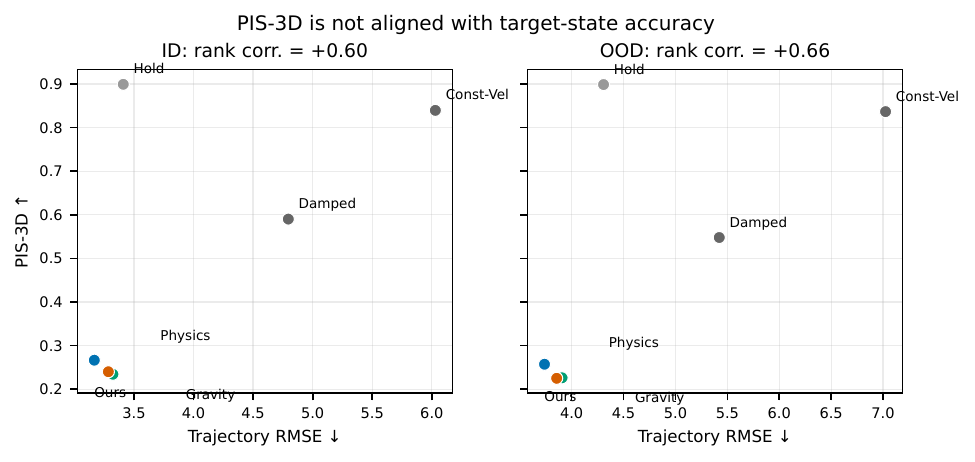}
\caption{PIS-3D versus trajectory RMSE for six non-oracle methods. The positive rank correlation is opposite to the sign expected of a score aligned with target-state accuracy.}
\label{fig:appendix-pis}
\end{figure*}

\FloatBarrier
Figure~\ref{fig:appendix-pis} compares PIS-3D with trajectory RMSE across the
six non-oracle methods. Their Spearman correlation is $+0.600$ on ID and
$+0.657$ on OOD. Since lower trajectory error should correspond to a higher
diagnostic score, an aligned ranking would have a negative sign.

\section{Qualitative Visualization Results}

This section presents the cross-method motion inventory, the pretrained
first-frame 3DGS-to-dynamics example, scene-aware background stress tests, and
an explicit analysis of the observed qualitative behavior.

\subsection{Cross-Method Motion Inventory}

\begin{figure*}[!ht]
\centering
\includegraphics[width=\textwidth]{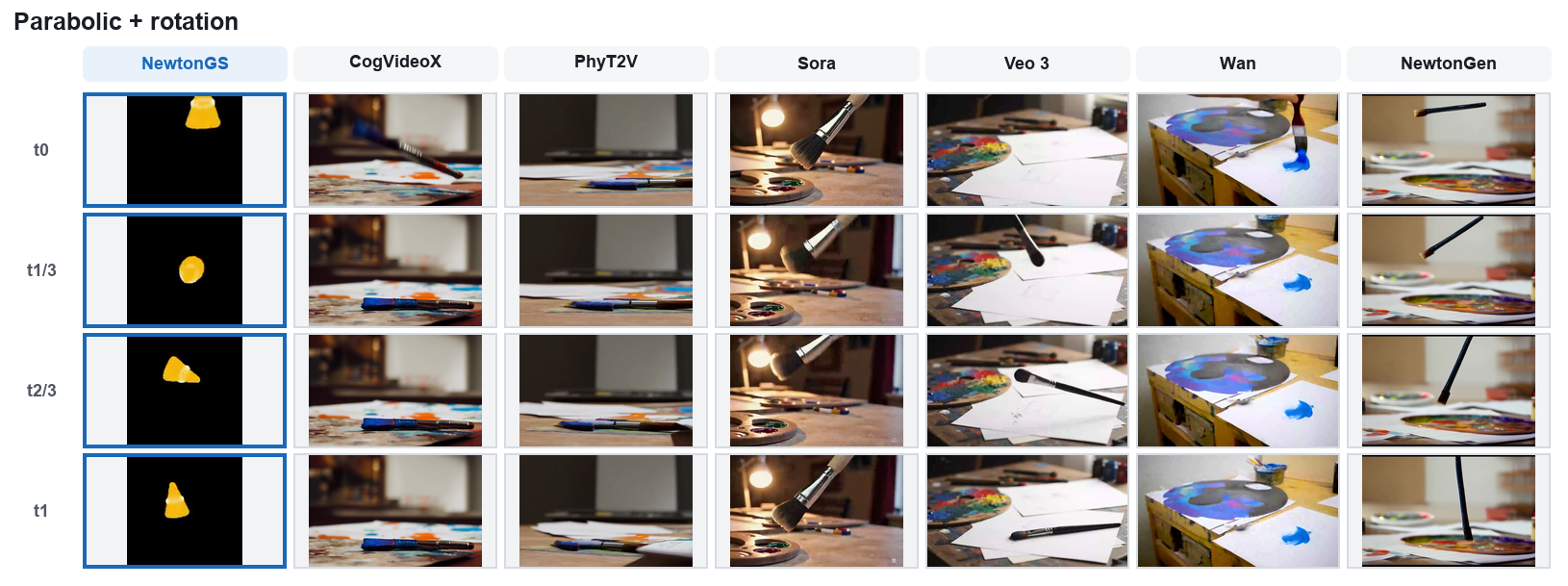}
\caption{Four-frame sequences for parabolic motion with rotation. The NewtonGS
column shows coupled displacement and orientation change while preserving the
Gaussian object's identity across the sampled times. All panels retain their
fixed-camera crops.}
\label{fig:appendix-video-parabolic-rotation}
\end{figure*}

Figure~\ref{fig:appendix-video-parabolic-rotation} retains the
intermediate-frame comparison for parabolic motion with rotation from the
public NewtonGen comparison set~\citep{yuan2026newtongen}. Each public-video
column is taken from the released example for that method, and the column
labeled NewtonGen corresponds to the release's \texttt{our.mp4}. NewtonGS uses
the fixed Same-32 proxy mapping and reference-camera view~0.

\paragraph{Panel construction.}
Blue borders identify NewtonGS. Each strip displays $t\in\{0,1/3,2/3,1\}$. The visualization script chooses frame index $\operatorname{round}(f(T-1))$, preserves aspect ratio with neutral padding, and applies no color, exposure, or content editing. Showing intermediate frames helps distinguish a visible state change from an isolated endpoint difference.
Here $f\in\{0,1/3,2/3,1\}$ is normalized sequence time, $T$ is the total
number of video frames, $T-1$ is the largest zero-based frame index, and
$\operatorname{round}$ selects the nearest integer index.

For NewtonGS, the fixed semantic mapping for this proxy category is tumbling
fall. The mapping is fixed before visual inspection.

\FloatBarrier

\begin{figure*}[!ht]
\centering
\begin{minipage}[t]{0.32\textwidth}
\centering
\textbf{Bouncing on plane}\\[2pt]
\includegraphics[
  width=\linewidth,
  trim=22.1bp 4.6bp 267.1bp 11.5bp,
  clip
]{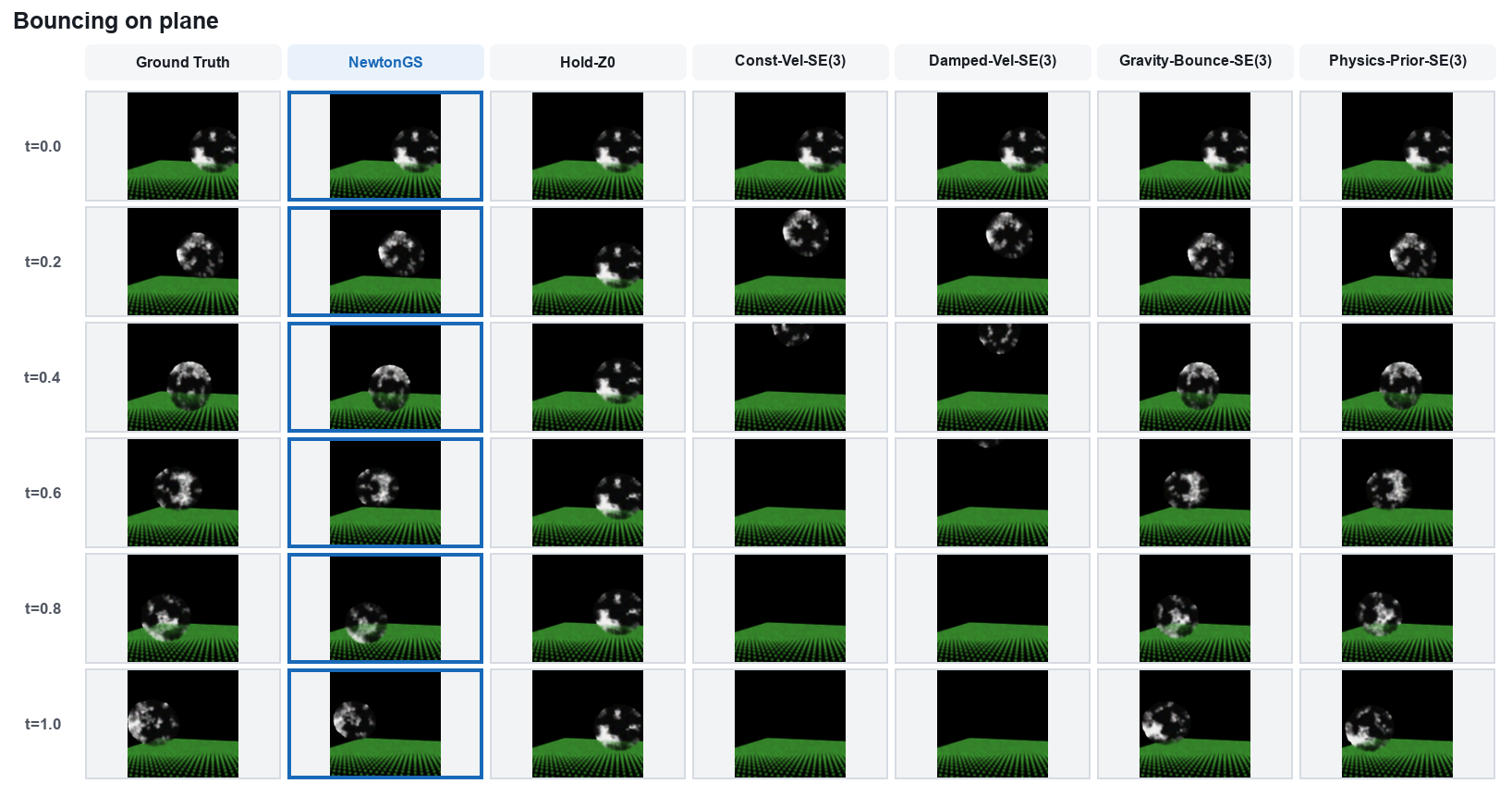}
\end{minipage}
\hfill
\begin{minipage}[t]{0.32\textwidth}
\centering
\textbf{Vertical launch}\\[2pt]
\includegraphics[
  width=\linewidth,
  trim=22.1bp 4.6bp 267.1bp 11.5bp,
  clip
]{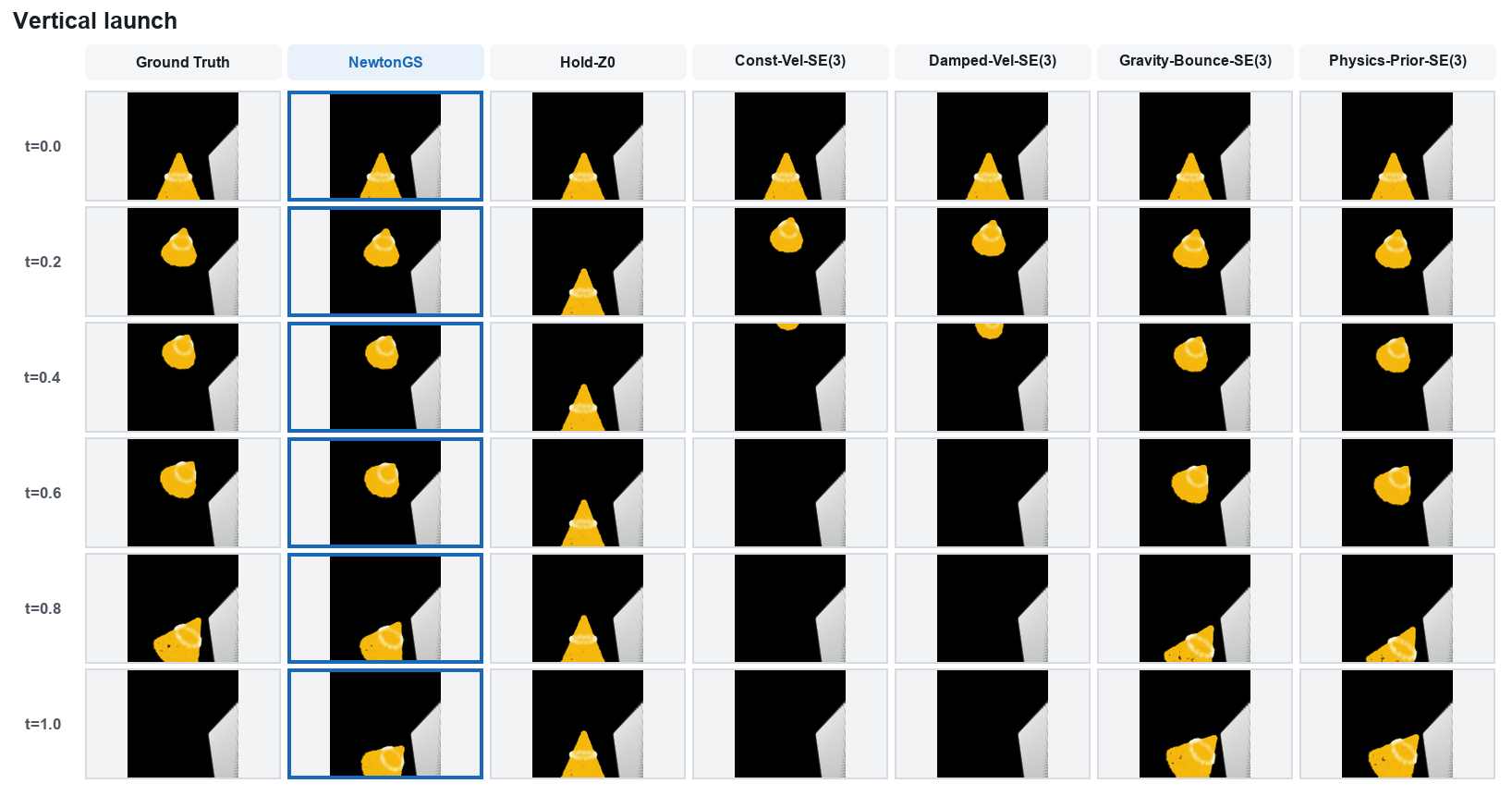}
\end{minipage}
\hfill
\begin{minipage}[t]{0.32\textwidth}
\centering
\textbf{Throw and land}\\[2pt]
\includegraphics[
  width=\linewidth,
  trim=22.1bp 4.6bp 267.1bp 11.5bp,
  clip
]{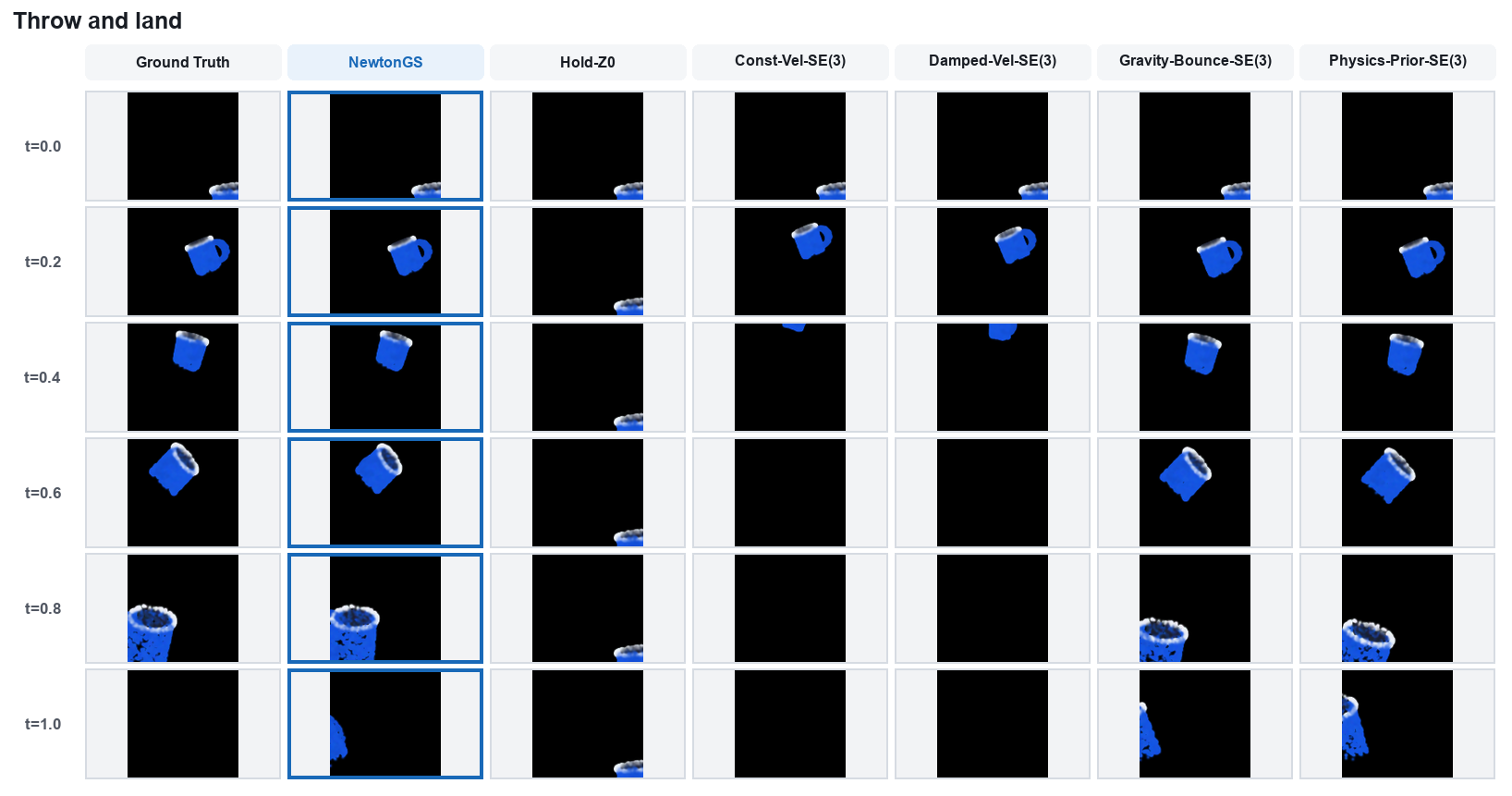}
\end{minipage}
\caption{Six-frame State-32 comparisons for bouncing on a plane,
vertical launch, and throwing followed by landing. Within each panel, ground
truth is shown on the left and NewtonGS on the right, with
$t\in\{0,0.2,0.4,0.6,0.8,1.0\}$ ordered from top to bottom. The source panels
are cropped only to remove the other baseline columns; no displayed frame is
altered or replaced.}
\label{fig:appendix-selected-newtongs}
\end{figure*}

\FloatBarrier
To complement the video-proxy comparison,
Figure~\ref{fig:appendix-selected-newtongs} shows three State-32 cases in which
the NewtonGS rollout remains visually close to the target over the displayed
sequence. The panels retain the complete temporal sampling from the original
method-comparison figures but show only the ground-truth and NewtonGS columns
for legibility.

\subsection{Pretrained First-Frame 3DGS Followed by Gaussian-NND}

We additionally instantiate the complete appearance-to-dynamics path requested in the main paper. A pretrained LGM front end~\citep{tang2024lgm} provides a single canonical 3DGS of a bowling ball in a complete bowling-alley scene. The asset contains 62,710 Gaussians, including the ball, lane, pins, walls, and ceiling lights. From the stored reference camera, a fixed circular image region and depth interval select 347 foreground Gaussians. To prevent the layered LGM representation from leaving a second static copy of the ball, 273 nonforeground Gaussians inside a smaller projected footprint are omitted; the remaining 62,090 scene Gaussians are static throughout the rollout. This projection-and-depth grouping is a documented visualization heuristic, not a learned segmentation result.

The pretrained 3DGS and its reference camera define the appearance and first-frame coordinate system. Gaussian-NND predicts a 49-frame object-state rollout. We convert state positions to displacements relative to $Z_0$, anchor those displacements at the selected foreground centroid, and apply the predicted relative rotation and scale to the foreground Gaussians. This coordinate conversion preserves the pretrained scene at $t=0$ while avoiding the incorrect assumption that the physical-state origin equals the LGM scene origin. The background is neither regenerated nor evolved.

For this qualitative probe, the initial physical input is chosen before
rendering by minimizing a fixed state-space score over 65,536 pseudorandom
candidates with seed 7302. The score targets a $+x$ displacement of 0.72 over
two seconds and penalizes vertical/lateral drift and scale change. This
selection is analogous to specifying a forward-motion physical prompt; it is
not a test-set result and is excluded from all quantitative tables.

\begin{figure*}[!ht]
\centering
\includegraphics[width=\textwidth]{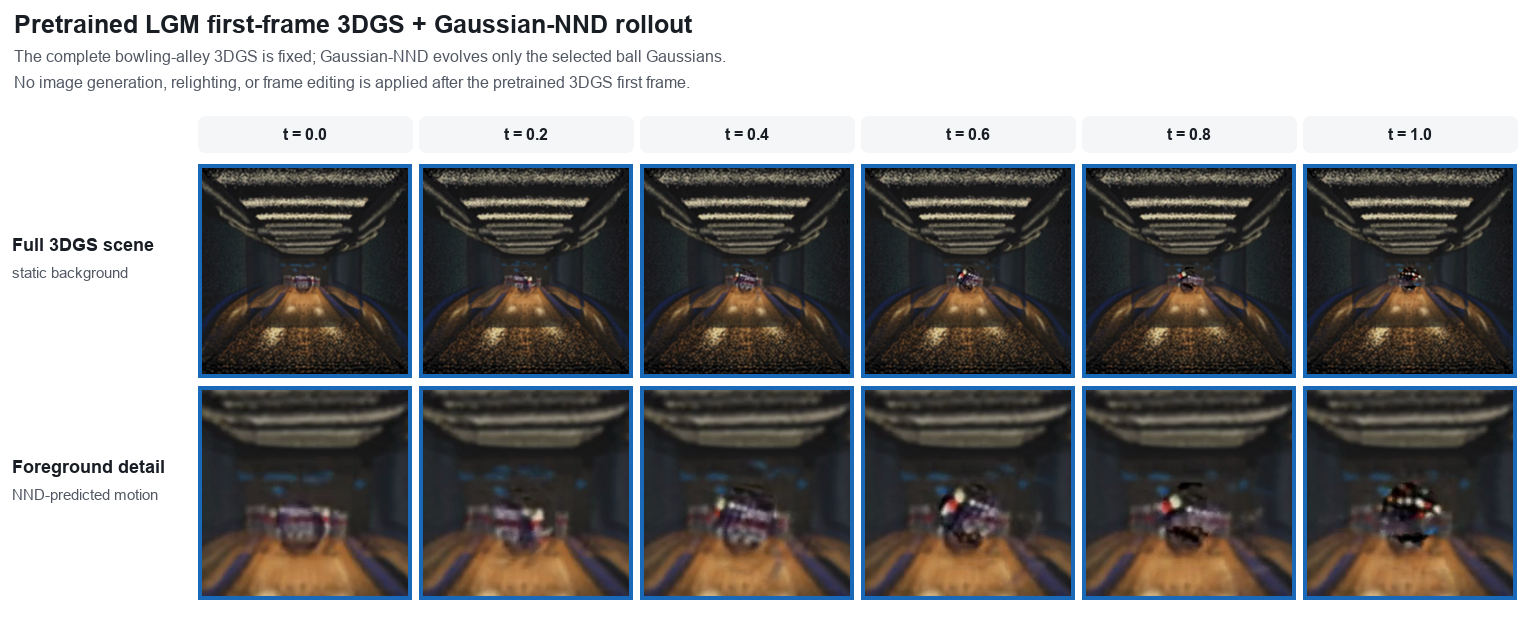}
\caption{Pretrained LGM~\citep{tang2024lgm} first-frame 3DGS followed by Gaussian-NND. The top row shows the complete static-background scene, and the bottom row magnifies the moving foreground region. Only selected ball Gaussians receive the predicted relative transform; no post-render frame editing is applied.}
\label{fig:appendix-pretrained-3dgs-nnd}
\end{figure*}

\FloatBarrier
Figure~\ref{fig:appendix-pretrained-3dgs-nnd} shows every uniformly spaced
displayed frame, including visible grouping and reconstruction artifacts.

\begin{figure*}[!ht]
\centering
\includegraphics[width=\textwidth]{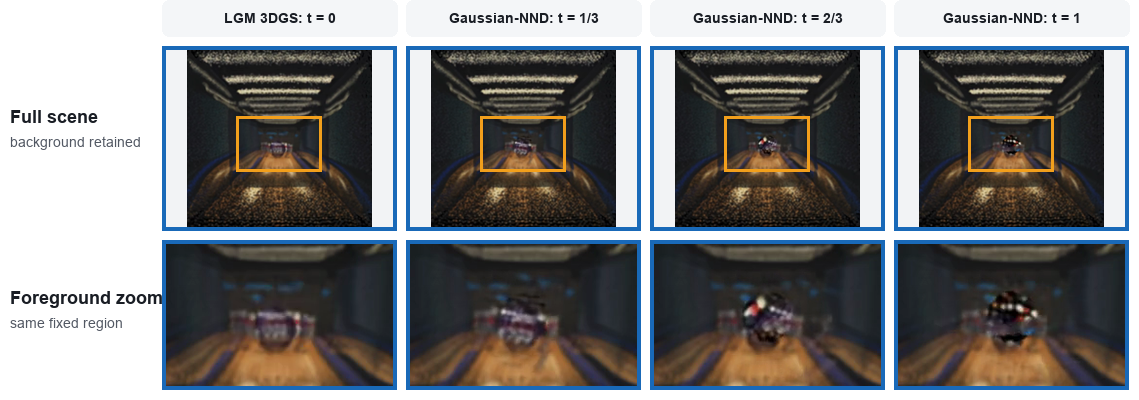}
\vspace{2pt}

\includegraphics[width=\textwidth]{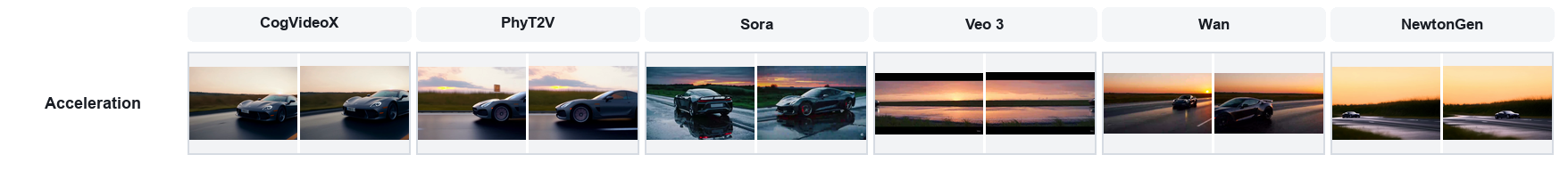}
\caption{Top: LGM~\citep{tang2024lgm} supplies the bowling-alley 3DGS at $t=0$, after which Gaussian-NND predicts foreground transforms while the background stays fixed; the second row magnifies the highlighted region. Bottom: exact-prompt public-video comparison from the NewtonGen release~\citep{yuan2026newtongen} (first/final frames). The shared prompt is ``A dark gray sports car accelerating in a straight line on an empty road at dusk, wet road reflecting light, orange sunset casting colors on the car, slight wind rustling the grass along the roadside, viewed from a stationary side-angle camera.'' NewtonGS is omitted from the bottom panel because it is conditioned on a Gaussian scene and physical state rather than text.}
\label{fig:appendix-video-examples}
\end{figure*}

\FloatBarrier
Figure~\ref{fig:appendix-video-examples} provides a compact juxtaposition of
this appearance-to-dynamics example and one exact-prompt public-video
comparison. The top panel is included to show object/background composition,
whereas the bottom panel documents the appearance variation among public
generators under one common text prompt.

\subsection{Procedural Scene-Aware Background Stress Tests}

The original NewtonGS comparison column uses a fixed construction-floor sample to keep the proxy mapping simple. To show the model in less uniform contexts, we additionally run the same selected checkpoint on 12 appearance-conditioned benchmark samples. Each sample contains a semantically matched colored object-centric Gaussian cluster and a static scene Gaussian set, such as a toy car on a living-room table, a soccer ball in a park, or a moon sphere in space. At inference, NewtonGS evolves only the object Gaussians. We then compose that predicted sequence with the sample's unchanged background Gaussians and render from the stored reference camera. Uncovered pixels use a scene-appropriate constant renderer color. No frame is retouched, inpainted, relit, or generated for presentation.

\begin{figure*}[!ht]
\centering
\includegraphics[width=\textwidth]{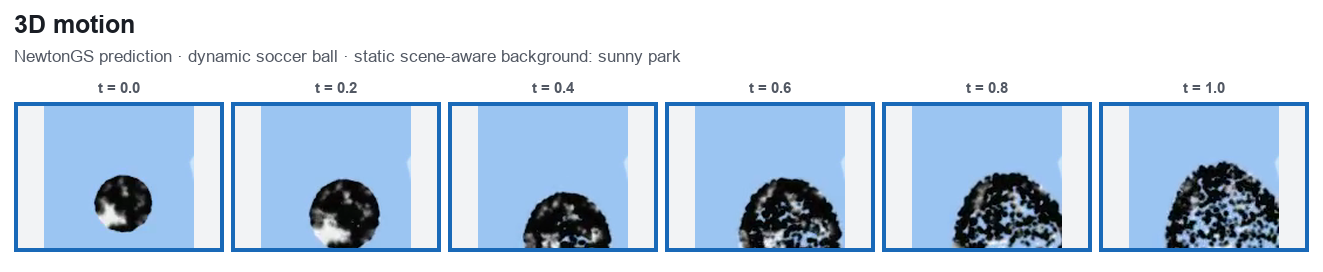}
\vspace{0.4em}
\includegraphics[width=\textwidth]{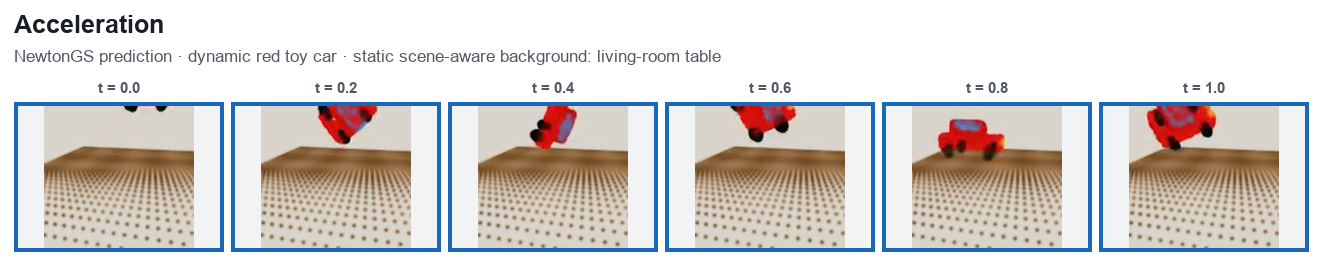}
\vspace{0.4em}
\includegraphics[width=\textwidth]{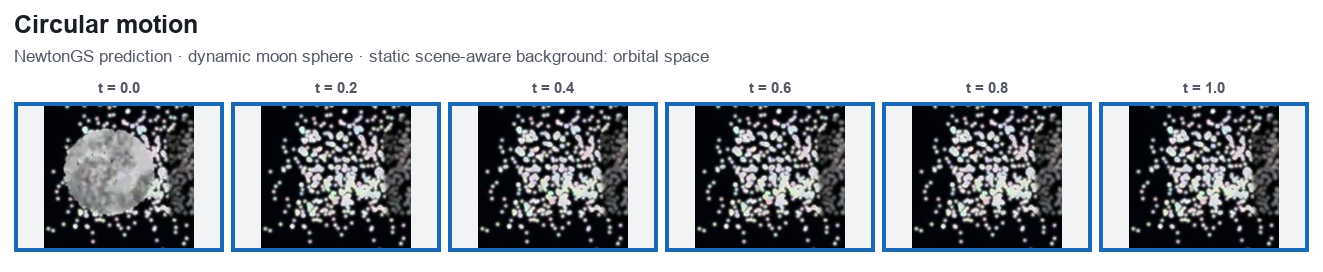}
\vspace{0.4em}
\includegraphics[width=\textwidth]{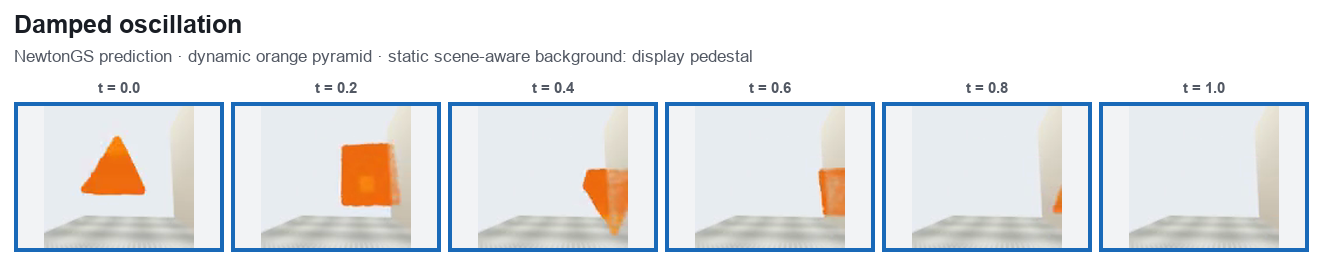}
\caption{Six-frame NewtonGS sequences with static scene backgrounds: 3D motion, acceleration, circular motion, and damped oscillation (top to bottom). The denser sampling exposes changes that can be hidden by endpoint-only display.}
\label{fig:appendix-rich-scene-a}
\end{figure*}

\FloatBarrier
Figure~\ref{fig:appendix-rich-scene-a} shows six uniformly spaced times for
3D motion, acceleration, circular motion, and damped oscillation. The static
background makes the predicted object displacement directly visible, while
the 3D-motion and circular-motion rows also expose loss of coherent Gaussian
extent.

\begin{figure*}[!ht]
\centering
\includegraphics[width=\textwidth]{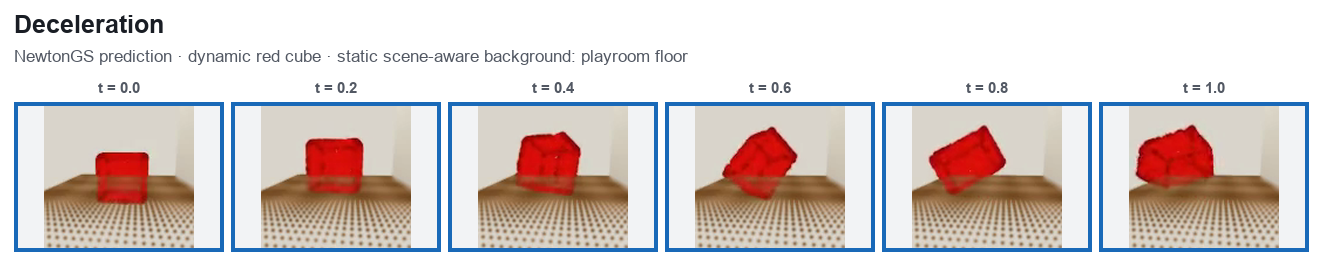}
\vspace{0.4em}
\includegraphics[width=\textwidth]{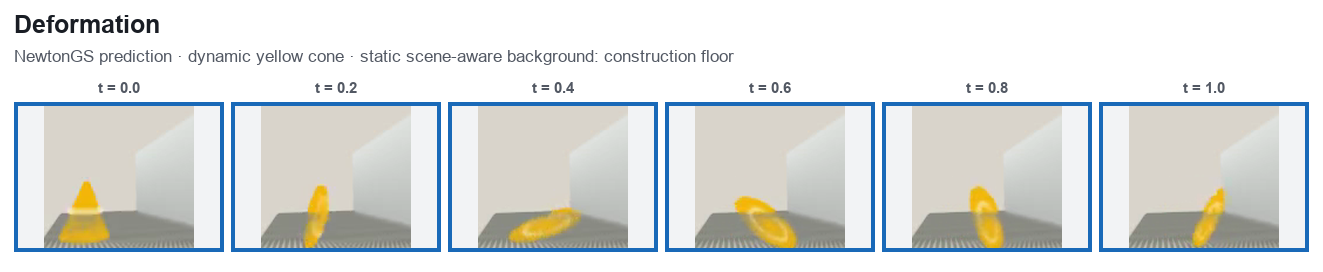}
\vspace{0.4em}
\includegraphics[width=\textwidth]{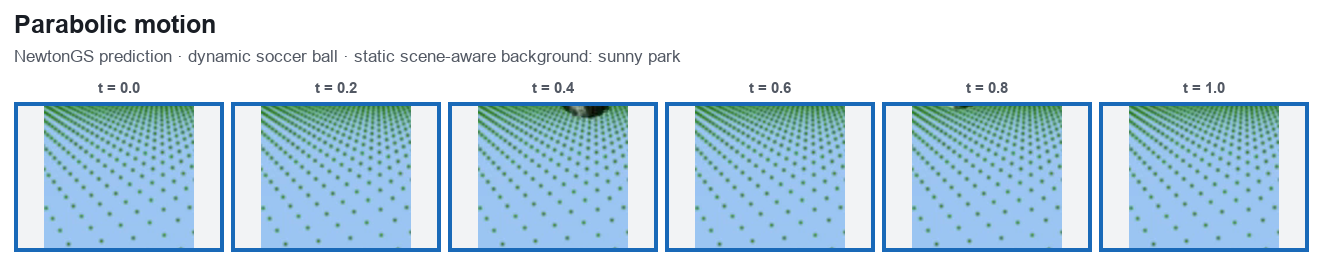}
\vspace{0.4em}
\includegraphics[width=\textwidth]{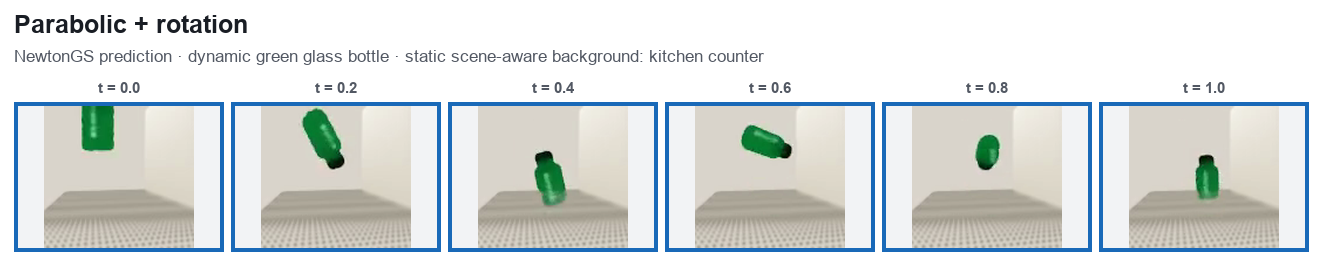}
\caption{Six-frame NewtonGS sequences with static scene backgrounds: deceleration, deformation, parabolic motion, and parabolic motion with rotation (top to bottom). Missing or clipped objects are shown without frame replacement.}
\label{fig:appendix-rich-scene-b}
\end{figure*}

\FloatBarrier
Figure~\ref{fig:appendix-rich-scene-b} shows deceleration, deformation,
parabolic motion, and parabolic motion with rotation. The panels preserve
missing or clipped objects and therefore expose both the requested motion
trend and the fixed-camera visibility failures.

\begin{figure*}[!ht]
\centering
\includegraphics[width=\textwidth]{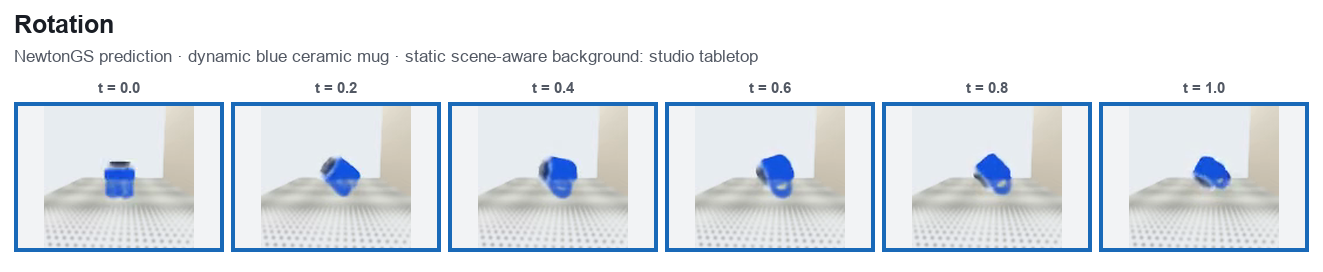}
\vspace{0.4em}
\includegraphics[width=\textwidth]{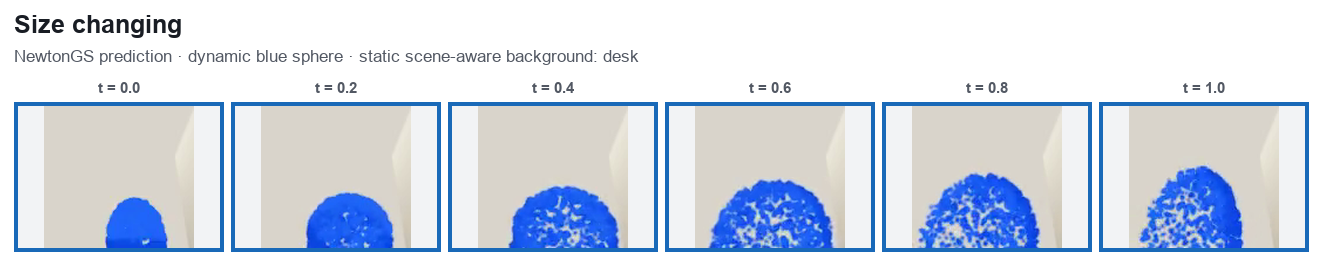}
\vspace{0.4em}
\includegraphics[width=\textwidth]{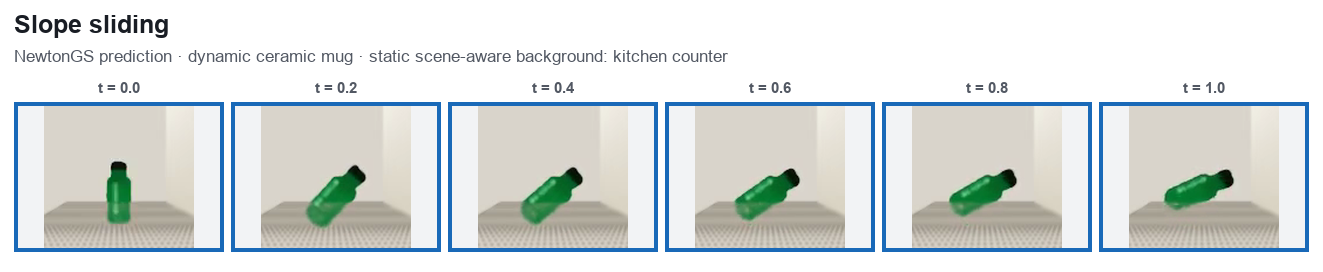}
\vspace{0.4em}
\includegraphics[width=\textwidth]{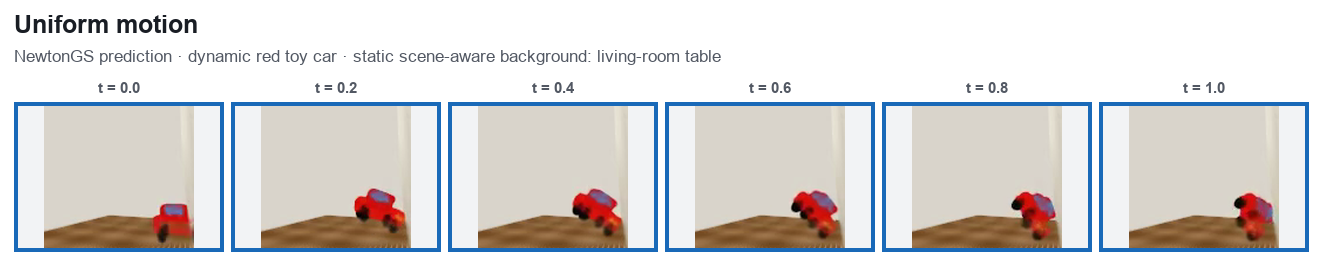}
\caption{Six-frame NewtonGS sequences with static scene backgrounds: rotation, size changing, slope sliding, and uniform motion (top to bottom). The background remains unchanged while the selected object Gaussians receive the predicted transform.}
\label{fig:appendix-rich-scene-c}
\end{figure*}

\FloatBarrier
Figure~\ref{fig:appendix-rich-scene-c} shows rotation, size changing, slope
sliding, and uniform motion. The backgrounds remain fixed while the object
Gaussians receive the predicted transform. Together with the preceding two
figures, these procedural scenes test the implemented object/background
composition path rather than learned background dynamics or photorealistic
generation.

\subsection{Qualitative Visualization Analysis}

\paragraph{Observed behavior and failure modes.}
In Figure~\ref{fig:appendix-video-parabolic-rotation}, the NewtonGS samples
visibly change position and orientation over time, so the state rollout is
reflected in the rendered object rather than only in an unrendered numerical
trajectory. The fixed reference camera can crop an object or let it leave the
field of view; this is a rendering-view limitation and should not be
interpreted as object deletion.

Figure~\ref{fig:appendix-selected-newtongs} presents three additional cases.
NewtonGS follows the alternating contact and airborne phases of the bouncing
sequence, the ascent and descent of the vertical launch, and the coupled
translation and rotation of the throw-and-land sequence.

The appearance-conditioned examples test a complementary part of the pipeline. Figure~\ref{fig:appendix-pretrained-3dgs-nnd} keeps the bowling-alley Gaussians fixed while the selected foreground subset moves through the six displayed times, demonstrating the implemented composition path from a pretrained first-frame 3DGS to an object-level rollout. The enlarged row also reveals speckling, blur, and an imperfect foreground grouping, so this example does not support a claim of reconstruction-quality preservation. In Figures~\ref{fig:appendix-rich-scene-a}-\ref{fig:appendix-rich-scene-c}, the static backgrounds remain temporally consistent and several objects exhibit the requested translation, rotation, or scale trend. At the same time, the 3D-motion, circular-motion, and size-changing rows show loss of coherent Gaussian extent, while some trajectories are partly outside the stored view. Taken together, the visualizations support the limited conclusion that the current implementation can transform selected object Gaussians and compose them with a static scene. They do not demonstrate photorealistic 4D reconstruction, local nonrigid dynamics, learned camera control, or superiority to text-to-video systems.

\section{Applicability, Failure Modes, and Observation Gap}

\paragraph{Contact and interaction scope.}
The event module detects only one object against the horizontal plane $y=0$.
It has no wall normal, inclined-plane normal, rolling constraint, pairwise
object broad phase, or multi-contact resolution. Wall collision, slope
sliding, rolling, and hybrid-contact labels in State-32 are therefore learned
only through the continuous residual and a floor event when that event happens
to trigger. This explains why wrong contact time, missing wall response, and
incorrect post-impact direction are expected failure modes rather than
violations of an implemented wall or slope solver.

\paragraph{Representation scope.}
Every selected object receives one shared relative rotation, translation, and
diagonal scale ratio. The method cannot represent articulated joints,
topology change, fracture, or a spatially varying deformation field.
``Non-rigid'' State-32 examples are only aggregate anisotropic-scale
trajectories. Covariance congruence can also exaggerate anisotropy when a
predicted scale component is inaccurate. Multiple objects can be stored in a
scene, but the reported rollout evolves one selected cluster at a time and
leaves every other Gaussian static.

\paragraph{Lifting failures.}
Opacity/object-weighted PCA is ill-conditioned when two eigenvalues are equal
or nearly equal. The code enforces a right-handed basis but does not perform
temporal axis-sign matching or eigenvector permutation tracking, so symmetric
objects can exhibit quaternion flips or unstable angular velocity. Soft mask
leakage changes the weighted center and extent; missing or fragmented masks can
move the state origin or split a coherent object. A single Gaussian frame
produces zero finite-difference linear, angular, and scale rates, and inserts
default material values; it cannot infer velocity, restitution, attenuation,
or mass from appearance.

\paragraph{Units and data domain.}
State-32 positions, velocities, and scales are synthetic generator units, not
calibrated meters, seconds-derived SI material parameters, or measured object
extents. The floor radius uses $|s_y|$ even though PCA lifting defines scale as
twice a weighted standard deviation. Gaussian-32 contains procedural objects,
procedural static backgrounds, known cameras, and labeled state trajectories;
it is not equivalent to reconstructing a real dynamic scene from images.

\paragraph{Observation-to-state gap.}
The headline State-32 results start from labeled $Z_0$, including velocities
and material controls. They do not evaluate automatic observation-to-state
inference. Projection grouping and PCA lifting are separate deterministic
interfaces used for Gaussian experiments and are not jointly trained with
Gaussian-NND. Consequently, State-32 accuracy is evidence about conditional
state rollout, while the bowling and scene-aware figures are interface
demonstrations. They do not establish end-to-end state estimation accuracy,
real-scene 4D reconstruction quality, or dynamic-background modeling.

\paragraph{View and presentation failures.}
The stored cameras are fixed and are not optimized to keep a predicted object
visible. Objects can leave the view, be cropped, or appear fragmented even
when the underlying state remains finite. Qualitative panels retain those
outcomes. Displayed temporal sequences use uniformly indexed normalized
times; rendering figures preserve aspect ratio with neutral padding and apply
no color, exposure, inpainting, or content correction.

\end{document}